\documentclass[sigconf=false, nonacm=true, review=false, anonymous = false, screen=true]{acmart}

\usepackage{graphicx}
\usepackage{multirow}
\usepackage{amsmath,amsfonts}
\usepackage{amsthm}
\usepackage{mathrsfs}
\usepackage{mathtools}
\usepackage[title]{appendix}
\usepackage{xcolor}
\usepackage{textcomp}
\usepackage{manyfoot}
\usepackage{booktabs}
\usepackage{algorithm}
\usepackage{algpseudocode}
\usepackage{listings}
\usepackage{xspace}
\usepackage{soul,xcolor}
\usepackage{bbm} 
\usepackage{amsbsy}
\usepackage{enumitem}
\usepackage{smartdiagram}
\usepackage{graphicx}
\usepackage{tikz}
\usepackage{subcaption}
\usepackage{tabularray}
\usepackage{float}
\usetikzlibrary{shapes,arrows,positioning}
\usepackage[stable]{footmisc}
\usepackage{adjustbox}
\usepackage{amsthm}
\usepackage{cleveref}

\newtheorem{theorem}{Theorem}
\newenvironment{delayedproof}[1]
 {\begin{proof}[\raisedtarget{#1}\textbf{Proof of \Cref{#1}}]}
 {\end{proof}}
\newcommand{\raisedtarget}[1]{%
  \raisebox{\fontcharht\font`P}[0pt][0pt]{\hypertarget{#1}{}}%
}
\newcommand{\proofref}[1]{\hyperlink{#1}{proof}}

\newcommand{\BF}[1]{
	\relax
	\ifmmode
	\ifcat\noexpand#1\relax %
		\boldsymbol{#1}     %
	\else
		\mathbf{#1}
	\fi
	\else
		\textbf{#1}
	\fi
}

\algdef{SE}[DOWHILE]{Do}{DoWhile}{\algorithmicdo}[1]{\algorithmicwhile\ #1}%

\DeclareMathOperator*{\argmin}{arg\,min}

\newcommand{\R}{\mathbb{R}}
\newcommand{\Z}{\mathbb{Z}}
\newcommand{\domain}{\ensuremath{\mathbb{X}}\xspace} %
\newcommand{\X}{\ensuremath{\mathbf{X}}\xspace} %
\newcommand{\F}{\ensuremath{F}\xspace} %
\newcommand{\Fest}{\ensuremath{\widehat{F}}\xspace} %
\renewcommand{\H}{\ensuremath{\mathbb{F}}\xspace}
\newcommand\inner[2]{\left\langle #1, #2 \right\rangle}
\newcommand{\norm}[1]{\left\lVert#1\right\rVert}
\newcommand{\set}[1]{\left\{#1\right\}}
\newcommand{\br}[1]{\left(#1\right)}
\newcommand{\cl}[1]{\left[#1\right]}
\renewcommand{\epsilon}{\varepsilon}
\newcommand{\eps}{\varepsilon}
\renewcommand{\mid}{\,\middle\vert\,}
\newcommand{\Span}[1]{\ensuremath{\operatorname{span}\set{#1}}} %

\theoremstyle{thmstyleone}

\theoremstyle{thmstyletwo}
\newtheorem{example}{Example}

\theoremstyle{thmstylethree}
\newtheorem{definition}{Definition}

\algrenewcommand\textproc{}%

\algnewcommand{\algorithmicand}{\textbf{ and }}
\algnewcommand{\algorithmicor}{\textbf{ or }}
\algnewcommand{\OR}{\algorithmicor}
\algnewcommand{\AND}{\algorithmicand}
\algnewcommand{\var}{\texttt}
\let\oldReturn\Return
\renewcommand{\Return}{\State\oldReturn}

\algnewcommand{\LineComment}[1]{\State \(\triangleright\) \textit{\color{blue} \scriptsize #1}}

\tikzset{
  treenode/.style = {align=center, inner sep=0pt, text centered,
    font=\sffamily},
  arn_n/.style = {treenode, circle, black, font=\sffamily, draw=black,
    fill=white, text width=1.5em},%
  arn_r/.style = {treenode, circle, white, draw=red, fill=red,
    text width=1.5em},%
  arn_b/.style = {treenode, circle, white, draw=blue, fill=blue,
    text width=1.5em},%
  arn_br/.style = {treenode, circle, black, draw=brown, fill=brown!40,
    text width=1.5em},%
  arn_v/.style = {treenode, circle, black, draw=blue, fill=violet!30,
    text width=1.5em},%
}

\tikzstyle{decision} = [diamond, draw, fill=orange!20, 
    text width=6em, text badly centered,node distance=2cm, inner sep=0pt]
\tikzstyle{block} = [rectangle, draw, fill=blue!20, 
    text width=6em, text centered, rounded corners, node distance=1.5 cm, minimum height=2em]
\tikzstyle{block2} = [rectangle, draw, fill=red!20, 
    text width=6em, text centered, rounded corners, node distance=1.5 cm, minimum height=2em]
\tikzstyle{line} = [draw, -latex']
\tikzstyle{cloud} = [draw, ellipse,fill=red!20, node distance=3cm,
    minimum height=2em]

\AtBeginDocument{%
  \providecommand\BibTeX{{%
    \normalfont B\kern-0.5em{\scshape i\kern-0.25em b}\kern-0.8em\TeX}}}

\setcopyright{rightsretained}
\copyrightyear{2024}
\acmYear{2024}
\acmDOI{XXXXXXX.XXXXXXX}
\acmConference[GECCO '24]{The Genetic and Evolutionary Computation Conference}{
  2024}{Melbourne, Australia}

\author{Kirill Antonov}
\affiliation{%
    \institution{LIACS, Leiden University}
    \streetaddress{Niels Bohrweg 1}
    \city{Leiden}
    \postcode{2333}
    \country{The Netherlands}}
\email{
k.antonov@liacs.leidenuniv.nl}
\orcid{0000-0002-8757-8598}

\author{Roman Kalkreuth}
\affiliation{%
    \institution{CNRS, LIP6, Sorbonne Universit\'e}
    \city{Paris}
    \country{France}}
\email{
roman.kalkreuth@lip6.fr}
\orcid{0000-0003-1449-5131}

\author{Kaifeng Yang}
\affiliation{%
    \institution{University of Applied Sciences Upper Austria}
    \city{Hagenger}
    \country{Austria}}
\email{
kaifeng.yang@fh-hagenberg.at}
\orcid{0000-0002-3353-3298}

\author{Thomas B{\"a}ck}
\affiliation{%
      \institution{LIACS, Leiden University}
      \streetaddress{Niels Bohrweg 1}
      \city{Leiden}
      \postcode{2333}
      \country{The Netherlands}}
\email{t.h.w.baeck@liacs.leidenuniv.nl}
\orcid{0000-0001-6768-1478}

\author{Niki van Stein}
\affiliation{%
    \institution{LIACS, Leiden University}
    \streetaddress{Niels Bohrweg 1}
    \city{Leiden}
    \postcode{2333}
    \country{The Netherlands}}
\email{n.van.stein@liacs.leidenuniv.nl}
\orcid{0000-0002-0013-7969}

\author{Anna V. Kononova}
\affiliation{%
    \institution{LIACS, Leiden University}
    \streetaddress{Niels Bohrweg 1}
    \city{Leiden}
    \postcode{2333}
    \country{The Netherlands}}
\email{a.kononova@liacs.leidenuniv.nl}
\orcid{0000-0002-4138-7024}

\begin{document}

\title{A Functional Analysis Approach to Symbolic Regression}

\begin{abstract}
Symbolic regression (SR) poses a significant challenge for randomized search heuristics due to its reliance on the synthesis of expressions for input-output mappings. 
Although traditional genetic programming (GP) algorithms have achieved success in various domains, they exhibit limited performance when tree-based representations are used for SR.
To address these limitations, we introduce a novel SR approach called Fourier Tree Growing (FTG) that draws insights from functional analysis. 
This new perspective enables us to perform optimization directly in a different space, thus avoiding intricate symbolic expressions. 
Our proposed algorithm exhibits significant performance improvements over traditional GP methods on a range of classical one-dimensional benchmarking problems. To identify and explain limiting factors of GP and FTG, we perform experiments on a large-scale polynomials benchmark with high-order polynomials up to degree 100.  
To the best of the authors' knowledge, this work represents the pioneering application of functional analysis in addressing SR problems.
The superior performance of the proposed algorithm and insights into the limitations of GP open the way for further advancing GP for SR and related areas of explainable machine learning. 
\end{abstract}

\maketitle

\section{Introduction}
Symbolic regression (SR) can be considered a major problem domain for search heuristics that focus on the synthesis of symbolic expressions. SR as a black-box optimization domain, aims at the derivation of mathematical expressions that are able to fit the input-output mapping of an unknown objective function to a predefined degree. 
In the wider domain of search heuristics, SR defines a diverse problem domain that has often been used to benchmark the performance of symbolic search algorithms but also offers a real-world application domain for such methods. Quite recently, it has been proven that SR is an NP-hard problem in view of the fact that it is not always possible to find the best-fitting mathematical expression for a given data set in polynomial time~\cite{DBLP:journals/tmlr/VirgolinP22}.
GP has been applied to SR problems since its early days. 
J. Koza~\cite{koza1990genetic, Koza:1992:GPP:138936, koza1994genetic} reported a series of results for the synthesis of symbolic expressions that can fit the functional behavior of polynomials of lower degrees by using a parse-tree representation model inspired by LISP s-expressions, which is now known as tree-based GP in the wider family of GP representations. Based on Koza's experiments, SR evolved to a major application domain for GP and was found to be the most popular problem in the first survey on benchmarking standards in GP at that time~\cite{DBLP:conf/gecco/McDermottWLMCVJKHJO12}. Besides the works that reported practical-oriented results by using GP, different aspects of GP have been analyzed since the early days of GP to understand its search behavior. Moreover, even if GP has been found to be a suitable method for SR, drawbacks and shortcomings that impede the effectiveness of the heuristic search in GP have been identified and studied in the past. Both, empirical and theoretical results, have been proposed to understand various properties of the evolutionary-inspired search mechanism, such as evolvability~\cite{kinnear:altenberg}, locality~\cite{10.1145/1830483.1830646}, fitness landscapes \& problem hardness~\cite{vanneschi:fca:gecco2004}, neutrality~\cite{Vanneschi2011, DBLP:conf/gptp/HuB16} and search \& runtime complexity~\cite{DBLP:conf/eurogp/MambriniO16,
DBLP:conf/aaai/LissovoiO18,DBLP:series/ncs/LissovoiO20}. Overall, works in these areas has made a significant contribution to the understanding of the heuristic search performed by GP. 

Despite recent advancements, the challenge of disruptive mutations and recombinations persists in SR, hindering the development of more efficient randomized search heuristics (RSH) in this domain.
In this paper, therefore, we take a step forward in the understanding of SR by considering the search from a general perspective of functional analysis (FA). %
We start by reformulating the SR problem as a classical norm-minimization problem in Hilbert Space. To solve this problem, we propose a novel method called Fourier Tree Growing (FTG) that is inspired by mutation-based GP as performed with a (1+$\lambda$) evolutionary algorithm (EA) and ramped half-and-half initialization that allows us to navigate in the considered Hilbert space.

Our experiments demonstrate that this method significantly outperforms the compared GP algorithms on classical one-dimensional benchmarking problems. %
We explain the observed gap in performance by considering the dynamics of the GP search process. %
To enable a detailed study, we propose a novel benchmark with high-order polynomials, which we call a large-scale polynomial benchmark (LSP).
We apply our proposed FTG method %
, conventional recombination-based GP as well as mutation-only GP to solve instances of LSP to identify and analyze shortcomings and limitations in the SR domain. 
We address the observed shortcomings and limitations of GP and FTG and discuss ways to overcome these drawbacks. Moreover, based on our theoretical and empirical findings, we discuss how some aspects of FA in general could be used in GP to benefit its application to SR. \\
The work presented in this paper aims at the following objectives:
\begin{itemize}
   \item Reformulation of the SR problem by means of FA
   \item Optimization in Hilbert space
   \item Evaluation of GP and FTG in the SR domain 
   \item Identification of aspects that could benefit GP in addressing SR problems
\end{itemize}
The results of this work can be easily adapted to canonical GP-based SR with a dimensionality reduction technique. %

The paper is structured as follows: In Section~\ref{sec:related_word} we briefly describe GP and SR and 
provide the preliminaries for the proposed FTG algorithm. %
In Section~\ref{sec:analysis_sr} we briefly formalize the corresponding SR problem statement and establish the framework for reformulation of the SR search problem %
as an optimization problem in Hilbert space. %
In Section~\ref{sec:hilbert_space} we propose FTG that performs SR search in Hibert space %
and rigorously study FTG's %
properties, and provide intuition on how it addresses SR. %
Section~\ref{sec:evaluation} is devoted to the presentation of empirical comparisons of the proposed algorithms against traditional GP search heuristics on conventional benchmarks and our proposed LSP benchmark. 
In Section~\ref{sec:discussion} we then discuss the results of our experiments. Finally, Section~\ref{sec:conclusion} concludes our work and outlines our future work.

\section{Related Work}
\label{sec:related_word}
\subsection{Genetic Programming}

Genetic Programming (GP) is a evolutionary-inspired search heuristic originally invented to enable the synthesis of computer programs for problem-solving. The main paradigm of GP is to evolve a population of \emph{computer programs} towards an algorithmic solution of a predefined problem. To accomplish this, GP transforms populations of candidate genetic programs, that are traditionally represented as parse-trees, iteratively from generation to generation into new populations of programs with (hopefully) better fitness. However, since GP is a stochastic optimization process, it can consequently not guarantee to achieve the optimal solution. \\
The first significant work in the field of GP was done by Forsyth~\cite{doi:10.1108/eb005587}, Cramer~\cite{cramer1985representation}, and Hicklin~\cite{Hicklin}. 
However, GP gained significantly more popularity when Koza applied his parse tree representation model, inspired by LISP S expressions, to several types of problems, for instance, symbolic regression, algorithm construction, logic synthesis, or classification~\cite{koza1990genetic,Koza:1992:GPP:138936,koza1994genetic}.
Besides the traditional tree-based representation model, GP variants with linear sequence representations~\cite{ieee94:perkis,Openshaw94buildingnew}, graph-based~\cite{Poli1996,miller:1999:ACGP} representations, or grammar-based representations~\cite{ryan:1998:geepal} have been proposed.
Traditional GP models variate candidate programs on a syntactical level while one of the most recently introduced GP models, Geometric Semantic GP, focuses on variation of candidate programs on a semantic level~\cite{DBLP:conf/ppsn/MoraglioKJ12}. \\
However, among the different forms of GP, tree-based GP can be considered the most popular representation model since it gained significant recognition in the evolutionary computation (EC) 
domain due to the experiments of Koza in problem domains that were practically relevant. However, despite its reputation in the field of EC for achieving practical results, GP suffers from drawbacks and shortcomings that have been found to hinder the effectiveness of the heuristic search. In the tree-based subdomain of GP, the most well-known drawback is \textit{Bloat} that is characterized by an uncontrolled growth of the average size of candidate trees in the population~\cite{10.5555/1595536.1595563}. Unlike most heuristic methods for numerical optimization in the EC domain, standard GP search operators such as subtree crossover and mutation provide only limited locality features~\cite{10.1145/1830483.1830646}, and have been found to be disruptive~\cite{DBLP:conf/eps/Angeline97}.

\subsection{Symbolic Regression in Genetic Programming}

Symbolic regression can be classified in the taxonomy of regression analysis, where a symbolic search on a space of mathematical functions is performed to find candidate functions that fit the ideal input-output mapping of a given dataset as close as possible, where the quality of the fit is typically measured on a given, finite set of data points.
Symbolic regression in GP can therefore be considered a black-box problem that forms a major problem domain in the application scope of GP since its very early days. In general, SR by means of GP relates to the application of GP models to synthesize mathematical expressions that represent the (unknown) function's input-output mapping as closely as possible. Symbolic regression gained prominence through Koza's pioneering efforts in the 1990s; however, the problem of finding a mathematical expression to explain empirical measurements was already introduced in the previous works~\cite{gerwin1974information,langley1981data,falkenhainer1986integrating}. In the early works of SR-based GP, Koza showed that GP can be used to discover SR models by encoding mathematical expressions as computational trees. Even though SR can be addressed by other algorithms (such as Monte Carlo tree search~\cite{cazenave2013monte,sun2022symbolic}, enumeration algorithms~\cite{kammerer2020symbolic}, greedy algorithms~\cite{de2018greedy}, mixed-integer nonlinear programming~\cite{cozad2018global}),  GP remains a popular choice. So far, SR through GP has been applied in different areas, such as economics~\cite{verstyuk2022machine}, medicine~\cite{virgolin2020machine}, engineering~\cite{kronberger2018predicting} and more~\cite{yang2023emo}. However, SR via GP still has some limitations, such as its gray-box property\cite{kotanchek2013symbolic}, model's over complexity~\cite{jackson2010identification}, and various models (with different structural properties, utilized variables)~\cite{affenzeller2014gaining}. 

\subsection{Preliminaries}
\label{subsec:prel}
Functional Analysis is a branch of mathematical analysis that studies functions, spaces of functions, and relationships between those spaces. %
One of the fruitful ideas used in FA is the notion of a Banach Space of which Hilbert Space is an important special case
We will briefly introduce this notion, and some relevant facts about it in this section because the notion of a Hilbert Space plays an important role in our work.

\begin{definition}[Metric Space] A \emph{metric space} is a pair $(X, d)$, where $X$ is a set and $d$ is a real-value function on $X \times X$ which satisfies that, for any $x,y,z \in X$, 
\begin{enumerate}
    \item $d(x,y) \geq 0$ and $d(x,y) = 0 \iff x = y$,
    \item $d(x,y) = d(y, x)$,
    \item $d(x,z) \leq d(x,y) + d(y,z)$.
\end{enumerate} 
The function $d$ is called the \emph{metric} on $X$. 
\label{def:n}
\end{definition} 

\begin{definition}[Normed Vector Space] A vector space $V$ over field $\mathbb{R}$ is called a \emph{normed vector space} if there is real-value function $\norm{\cdot}$ on $V$, called the \emph{norm}, such that for any $x,y \in V$ and any $\alpha \in \mathbb{R}$,
\begin{enumerate}
    \item $\norm{x} \geq 0$ 
    \item $\norm{x}=0 \iff x = 0$,
    \item $\norm{\alpha x} = |\alpha|\norm{x}$,
    \item $\norm{x+y} \leq \norm{x} + \norm{y}$.
\end{enumerate}
\end{definition}
\begin{definition}[Cauchy Sequence]
\label{def:cs}
A sequence $\set{x_n}$ in a metric space $(X, d)$ is a \emph{Cauchy Sequence} if
$$\forall \epsilon > 0, \exists N \in \mathbb{N}: d(x_n, x_m) < \epsilon \quad \forall n,m > N.$$
\end{definition} 
\begin{definition}[Complete Space]
    A normed space is called complete if every Cauchy sequence in the space converges to an element from this space.
\end{definition}
\begin{definition}[Inner Product Space]
\label{def:ips}
An \emph{inner product space} is a vector space $V$ over the field $\mathbb{R}$ together with an inner product, that is a map: 
$$\inner{\cdot}{\cdot}_V : V \times V \rightarrow \mathbb{R}$$ with the following properties:
\begin{enumerate}
    \item $\inner{\alpha(x+y)}{z}_V=\alpha\inner{x}{z}_V+\alpha\inner{y}{z}_V$,
    \item $\inner{x}{y}_V = \inner{y}{x}_V$,
    \item $\inner{x}{x}_V \ge 0$,
    \item $\inner{x}{x}_V = 0 \iff x = 0_V$.
\end{enumerate}
\end{definition}

\begin{definition}[Hilbert Space]
\label{def:hs}
A complete inner product space is a Hilbert space.
\end{definition}

\begin{theorem}[Projection Theorem \cite{kantorovich2016functional}]
    Consider Hilbert space $\mathcal{X}$ and its complete linear subspace $\mathcal{Y}$.
    \begin{enumerate}
        \item For all elements $x \in \mathcal{X}$, there exists a unique element $y^* \in \mathcal{Y}$ such that \[\norm{x - y^*} = \min\limits_{z \in \mathcal{Y}} \norm{x - z}\]
        This element $y^*$ is called the \textit{closest} to $x$ in the subspace $\mathcal{Y}$.
        \item Given the closest element $y^*$ to $x$ in the subspace $\mathcal{Y}$, for all $x \in \mathcal{X}$ 
        and $z \in \mathcal{Y}$, we have $\inner{x - y^*}{z} = 0$.
    \end{enumerate}
    \label{th:prj}
\end{theorem}

\begin{definition}[Linear Independence]
    Elements $\set{\BF{v}_1, \BF{v}_2, \dots, \BF{v}_k}$ of a vector space over the field $\R$ are linearly independent if for every set of constants $\set{\alpha_1, \alpha_2, \dots, \alpha_k}$ either $\sum_i^k \alpha_i \BF{v}_i \ne \BF{0}$ or $\forall i: \alpha_i = 0$. 
    Otherwise, they are linearly dependent.
    \label{def:ld}
\end{definition} 

Consider linear independent elements from Hilbert space $\mathcal{H}$: $h_1, h_2, \dots, h_k$ and element $g \in \mathcal{H}$. %
We can formulate a \textit{linear least squares approximation problem} (LLSQ) in this Hilbert space as finding such constants $\alpha_1, \alpha_2, \dots, \alpha_k$ that
$\norm{g - \alpha_1h_1 - \alpha_2h_2 - \dots - \alpha_kh_k}$ is minimized.
Due to Theorem \ref{th:prj}, there exists a unique element $h^* \in \Span{h_1, h_2, \dots, h_k}$ closest to $g$.
It is clear, that this $h^*$ is the solution of the stated LLSQ.
Due to the second statement of Theorem \ref{th:prj}, all of the following equalities hold simultaneously: \[\inner{g-h^*}{h_1}_\mathcal{H}=0, \inner{g-h^*}{h_2}_\mathcal{H}=0, \dots, \inner{g-h^*}{h_k}_\mathcal{H}=0\]
It is equivalent to the following matrix equation: \begin{equation}\br{\inner{g}{h_1}_\mathcal{H}, \inner{g}{h_2}_\mathcal{H}, \dots, \inner{g}{h_k}_\mathcal{H}}^\top = \BF{G}\cdot\br{\alpha_1, \alpha_2, \dots, \alpha_k}^\top  \label{eq:matrix-eq}\end{equation}
where $1 \le i,j \le k$ and $\BF{G} = \br{\inner{h_i}{h_j}_\mathcal{H}}$ is the \textit{Gram matrix}.

For arbitrary $\BF{\alpha} \in \R^k$ we have 
\begin{align*}
    \BF{\alpha}^\top\BF{G}\BF{\alpha} 
    & = \sum\limits_{1 \le i,j \le k}{\alpha_i\alpha_j\inner{h_i}{h_j}_\mathcal{H}}  \\
    & = \inner{\sum\limits_{1 \le i \le k}\alpha_ih_i}{\sum\limits_{1 \le i \le k}\alpha_ih_i}_\mathcal{H}  \\
    & = \norm{{\sum\limits_{1 \le i \le k}\alpha_ih_i}}^2 \ge 0
\end{align*}
The equality is attained for $\BF{\alpha} \ne 0_{\R^k}$ if and only if elements $h_1$, $h_2$, $\dots$, $h_k$ are linearly dependent.
So, the Gram matrix is positive definite for the considered linearly independent elements.
It implies that all its eigenvalues $\lambda_1, \lambda_2, \dots, \lambda_k$ %
are real positive constants, so $\det(\BF{G}) = \lambda_1\cdot\lambda_2\cdot\dotsc\cdot\lambda_k > 0$.
Then its inverse exists, and we obtain the following solution of the matrix equation Eq. \eqref{eq:matrix-eq}:
\begin{equation}
     \br{\alpha_1, \alpha_2, \dots, \alpha_k}^\top = \BF{G}^{-1}\cdot\br{\inner{g}{h_1}_\mathcal{H}, \inner{g}{h_2}_\mathcal{H}, \dots, \inner{g}{h_k}_\mathcal{H}}^\top
     \label{eq:prj_to_subspace}
\end{equation}

\section{Analysis of Symbolic Regression in the General Case}
\label{sec:analysis_sr}

Building upon the concepts introduced in Sec.~\ref{subsec:prel}, this section reframes the optimization of the SR problem through functional analysis.
We start by formulating the conventional SR problem more formally.

\subsection{Problem Formalization}

We formalize Symbolic Regression in the following form.

Let us consider:
\begin{enumerate}
    \item Domain $\domain \coloneqq \prod \limits_{i = 1}^n[a_i, b_i] \subset \R^n$;
    \item An unknown arbitrary function $\F : \domain \to \R$ is to be approximated. We refer to this function $F$ as \textit{target} function;
    \item Ordered set of points $\X \coloneqq \set{\BF{x}_1, \BF{x}_2, \dots, \BF{x}_N} \subset \domain$ 
    , which we consider as \textit{training data}. %
    It is used as the set of points where the values of the function \F are known; 
    \item Unary operators $\mathfrak{U} \subseteq \left\{u \mid u : \R \to \R \right\}$ %
    , binary operators $\mathfrak{B} \subseteq \left\{b \mid b: \R^2 \to \R\right\}$, constants $\mathfrak{C} \subseteq \R$ and orthogonal projection operators $\Pi \coloneqq \left\{(p_i)_{i=1}^n\right\}$, $p_i: \domain \to \R$, $p_i(x) = (0,\dots,1,0,\dots)\cdot x$, where $1$ is in the $i$-th position. 
    We will refer to them as \textit{elementary} operators.
\end{enumerate}

The goal of Symbolic Regression is to obtain an \emph{estimated}
function $\Fest : \domain \to \R$ such that:
\begin{enumerate}
    \item $\Fest$ is a composition of operators from $\mathfrak{U} \cup \mathfrak{B} \cup \Pi$; \label{goal:composition}
    \item 
    \label{goal:minimizer}
    $\Fest$ is a minimizer of the following functional  (\emph{loss function):} 
    \begin{equation*}
        \mathcal{L}\br{\Fest} \coloneqq \frac{1}{N} \cdot 
        \sum_{i=1}^N
        {f_{itness} \br{  \F(\BF{x}_i), \Fest(\BF{x}_i)}}\; ,
    \end{equation*}
    where $f_{itness}(.)$ is the fitness function in GP. Commonly, $f_{itness}(.)$ can be choosen as MSE, NMSE, R2.
    In this paper, we choose $f_{itness}(.)$ as squared error (SE), that is $f_{itness}(x,y) \\ \coloneqq (x-y)^2$.
    \item Among all such estimated %
    functions that satisfy goals \eqref{goal:composition} and \eqref{goal:minimizer} %
    , \Fest has the smallest length of the description.\label{goal:kolmogorov} %
\end{enumerate}

GP algorithms that address the formulated problem in practice are usually limited in terms of resources and so can not afford to work infinitely long. 
Conventionally, \emph{the number of traverses} over the dataset \X is considered as an indication of time spent on solving the problem.
For example, the computation of the loss function costs one time unit.
In this work, \emph{the number of traverses} is used as well to analyze the algorithms' performance on generating a sufficiently accurate function \Fest. 

\subsection{Problem Reformulation %
}

Let us consider a vector space of all functions $\mathcal{F} \coloneqq \set{f : \domain \to \R}$.
Since we are working with the fixed set of training data $\X$, two different functions in $\mathcal{F}$ are seen as the same function when an instance of SR %
is approached.
This motivates the following relation for the set $\mathcal{F}$.
\begin{definition}
    \label{def:eq}
    Functions $f,g : \domain \to \R$ are called $\X$-identical, $f \sim g$ if $~\forall \BF{x} \in \X: f(\BF{x}) = g(\BF{x})$
\end{definition}
It is easy to see that this relation $\sim$ is an equivalence relation. 
Let us denote the set of all equivalence classes of functions $\domain \to \R$ as \H.
By construction, we obtained the \emph{quotient space} $\H = \mathcal{F}/\sim$.
To not mix up functions with classes of equivalent functions, we will use the notation $\cl{\cdot}$
, which is defined as $\cl{f} \coloneqq \set{g : \domain \to \R \mid f \sim g}$.
Now we propose Theorem \ref{th:hilbertspace} that collects several statements about the set \H and introduces an inner product, which plays a crucial role in this work.

\begin{theorem}
\label{th:hilbertspace}
    \H is a vector space over field $\R$ when equipped with:\\
    \begin{tabular}{p{0.01\textwidth} p{0.13\textwidth} l}
        (1) & operator $+:$ & $\cl{f} + \cl{g} = \cl{\forall \BF{x} \in \domain: \BF{x} \mapsto f(\BF{x}) + g(\BF{x})}$ \\
        (2) & scalar mult.: & $r\cl{f} = \cl{\forall \BF{x} \in \domain: \BF{x} \mapsto rf(\BF{x})} $ \\
        (3) & neutral element: & $ \BF{0} = \cl{\forall \BF{x} \in \domain: \BF{x} \mapsto 0}$ \\
    \end{tabular}
    Moreover, this vector space \H is a $N$-dimensional Hilbert space with inner product:
    \[\inner{\cl{f}}{\cl{g}} \coloneqq \sum\limits_{\BF{x} \in \X} f(\BF{x})g(\BF{x})\]
\end{theorem}
See \proofref{th:hilbertspace} in the appendix.

The defined inner product induces the following norm and distance in Hilbert space \H:
\begin{gather}
\norm{\cl{f}} = \sqrt{\sum\limits_{\BF{x} \in \X} f^2(\BF{x})} \label{eq:norm} \\
d\br{\cl{f}, \cl{g}} = \sqrt{\sum\limits_{\BF{x} \in \X} \br{f(\BF{x})-g(\BF{x})}^2} \label{eq:dist} 
\end{gather}

It is convenient to consider one particular distance between an arbitrary element $\cl{f} \in \H$ and the equivalence class which contains the target function \F, i.e., 
$d^2\br{\cl{f}, \cl{\F}} = \mathcal{L}(f)$.
Now, we can transform the goal \ref{goal:minimizer} listed before to an optimization problem in Hilbert space \H: 
\begin{equation}
\Fest^* = \argmin\limits_{\Fest}d\br{\cl{\Fest}, \cl{F}}
\label{eq:opt}
\end{equation}

In the following section, we propose an approach to solve this optimization problem by generating a solution as a composition of elementary operators. 
However, we do not directly address the goal \ref{goal:kolmogorov} defined before and mostly focus only on the accuracy of the obtained functions.
Minimizing the length of the produced expression is left for future works.

\section{Optimization in Hilbert Space}
\label{sec:hilbert_space}
Depending on the chosen elementary operators, some functions in \H might be not representable by a finite composition of elementary operators.
This implies that achieving the global minimum value of zero for the optimization problem in Eq.~\eqref{eq:opt} may not be reachable.
In practical cases, it is relevant to support a rich %
set representable functions, so it is reasonable to assume that the set of elementary operators is big.
Hence, we assume that basic operators are included in elementary operators.
Particularly, we consider that addition and multiplication belong to the binary operators $\mathfrak{B}$, and all real constants $\R$ belong to the set $\mathfrak{C}$. 
Given these minimalistic assumptions on elementary operators, we look for a minimizer of Eq.~\eqref{eq:opt} in the form: 
\[\Fest: \BF{x} \mapsto \alpha_1 v_1(\BF{x}) + \alpha_2 v_2(\BF{x}) + \dots + \alpha_k v_k(\BF{x}),\]
where functions $v_1, v_2, \dots, v_k$ are some compositions of elementary operators, that is $v_i \in \mathfrak{U} \cup \mathfrak{B} \cup \Pi \cup \mathfrak{C}, i \in \Z_k$. %

Let us assume that elements $\cl{v_1}$, $\cl{v_2}$, $\dots$, $\cl{v_k}$ are linearly independent. The linear combination $\sum_i^k{\alpha_i v_i(\BF{x})}$ of elements from $\mathcal{F}$ still belongs to the space $\mathcal{F}$, so we can consider element $\cl{\sum_i^k{\alpha_i v_i(\BF{x})}}$ of space \H. Based on the properties of quotient space, $\cl{\sum_i^k{\alpha_i v_i(\BF{x})}} = \sum_i^k \alpha_i \cl{v_i(\BF{x})}$. Since all the constants are allowed, elements $\cl{v_1}$, $\cl{v_2}$, $\dots$, $\cl{v_k}$ span a complete subspace of quotient %
space \H. From the projection theorem and Eq. \eqref{eq:prj_to_subspace}, we obtain that in this case, there exists a single closest element in this subspace to the element $\F$. However, when functions $v_1$, $v_2$, $\dots$, $v_k$ are linearly dependent, the determinant of the corresponding Gram matrix is zero, and so the coefficients $\alpha_i, i=1,\dots,k$, which gives the closest point to $\F$, can not be computed. 

We propose the following Algorithm \ref{alg:ftg} as a general heuristic algorithm for solving SR instances. 
In Theorem \ref{th:hilbertspace} we showed that space \H is finite-dimensional, however, the following algorithm does not require this property.
In this regard, we formulate the algorithm in the general case of Hilbert space, which might be infinitely dimensional.
The main idea of the algorithm is to generate compositions of elementary operators and ensure that all of the generated functions %
are linearly independent. 
It allows us to apply the projection theorem and get the best possible approximation of $\F$ in the subspace spanned by the functions.

\begin{algorithm}[]
\caption{General scheme of Fourier Tree Growing (FTG) algorithm in Hilbert space $\mathcal{H}$} 
\begin{algorithmic}[1]
    \State{$v_1 \gets \br{\forall \BF{x} \in \domain: \BF{x} \mapsto 1}$} \label{alg:ftg:constant-1}
    \State{$\Fest_1 \gets \br{\forall \BF{x} \in \domain: \BF{x} \mapsto \inner{\cl{v_1}}{\cl{v_1}}^{-1} \inner{\cl{\F}}{\cl{v_1}}}$} \label{alg:ftg:Fest_1}
    \For{$k \gets 2, 3, \dots$} \label{alg:ftg:loop}
        \State{\textbf{if} $\mathcal{L}(\widehat{\F}_{k-1}) = 0$ \textbf{then} \textbf{ return } $\widehat{F}_{k-1}$ \textbf{end if} }\label{alg:ftg:ret1}
        \Do
            \State{$v_k \gets \textsc{generate-composition}\br{\mathfrak{U}, \mathfrak{B},  \mathfrak{C}, \Pi, p, l, u}$} \label{alg:ftg:gen-comp} %
        \DoWhile{$\inner{\cl{\F}-\cl{\widehat{F}_{k-1}}}{\cl{v_k}} = 0$} \label{alg:ftg:condition}

        \State{$\BF{G} \gets 
        \begin{pmatrix}
            \inner{\cl{v_1}}{\cl{v_1}} & \inner{\cl{v_1}}{\cl{v_2}} & \dots & \inner{\cl{v_1}}{\cl{v_k}} \\
            \inner{\cl{v_2}}{\cl{v_1}} & \inner{\cl{v_2}}{\cl{v_2}} & \dots & \inner{\cl{v_2}}{\cl{v_k}} \\
            \vdots & \vdots & \ddots & \vdots \\
            \inner{\cl{v_k}}{\cl{v_1}} & \inner{\cl{v_k}}{\cl{v_2}} & \dots & \inner{\cl{v_k}}{\cl{v_k}}
        \end{pmatrix}
        $}\label{alg:ftg:gram}
        \State{$\BF{\alpha} \gets \BF{G}^{-1}
        \begin{pmatrix}
            \inner{\cl{F}}{\cl{v_1}} \\
            \inner{\cl{F}}{\cl{v_2}} \\
            \vdots \\
            \inner{\cl{F}}{\cl{v_k}}
        \end{pmatrix}$}\label{alg:ftg:alpha}
        \State{$\Fest_k \gets \br{\forall \BF{x} \in \domain: \BF{x} \mapsto \alpha_1 v_1(\BF{x}) + \alpha_2 v_2(\BF{x}) + \dots + \alpha_k v_k(\BF{x})}$} \label{alg:ftg:update}
    \EndFor
    \Return{$\widehat{F}_k$}
\end{algorithmic}
\label{alg:ftg}
\end{algorithm}

Algorithm \ref{alg:ftg} iteratively generates functions $v_k$ and uses their linear combination as $\Fest$.
The algorithm uses the Gram matrix in line \ref{alg:ftg:gram} to obtain the coefficients $\BF{\alpha}$ for the linear combination.
If the Gram matrix is the identity, then elements of $\BF{\alpha}$ are called Fourier coefficients.
In our case, the Gram matrix is almost never identity, but we still use the word Fourier to refer to those coefficients.
Since GP uses trees to represent the composition of functions, the tree that represents the obtained $\Fest$ contains coefficients $\BF{\alpha}$ in some nodes.
This is the reason to call the algorithm Fourier Tree Growing.

The generation of the composition of function in line \ref{alg:ftg:gen-comp} is performed using a heuristic algorithm, which samples elementary operators from a parameterized probability distribution, and creates their compositions.
This heuristic and the distribution are discussed later in Section \ref{sec:impl}.

We can consider the generation of the function $v_k$ as the application of mutation operator to the current approximation $\Fest_{k-1}$ of the target function \F.
Using an evolutionary metaphor, we can define $\Fest_k$ as the $k$-th candidate solution.
It turns out, that the mutation which violates the predicate in line \ref{alg:ftg:condition} increases the quality of the produced individual $\Fest_k$, which is shown rigorously in Theorem \ref{theorem:4opt}.
In this case, Algorithm \ref{alg:ftg} can be seen as an adaptation of a $(1 + 1)$-EA for optimization in Hilbert space.

\begin{theorem}
    Consider sequence of functions $(\Fest_k)_{k=1}^N$ constructed by Algorithm \ref{alg:ftg}. For every $i > j: \mathcal{L}(\Fest_i) < \mathcal{L}(\Fest_j)$. 
    Moreover, if $\mathcal{H}$ is finite dimensional with $N$ dimensions, then algorithm returns function $\Fest$ such that $\mathcal{L}(\Fest) = 0$. %
\label{theorem:4opt}
\end{theorem}
See \proofref{theorem:4opt} in the appendix.

Due to the unordinary definition of the time that the algorithm takes to solve an SR instance, we say explicitly where the algorithm spends the computational budget.
We defined that the algorithm spends one time unit when the set of training data \X is traversed one time.
This traverse happens when the inner product is computed or the loss function is computed.
Computation of the inner product is performed twice in the line \ref{alg:ftg:Fest_1}.
On every iteration of the loop in line \ref{alg:ftg:loop} the FTG algorithm computes the value of loss directly in line \ref{alg:ftg:ret1} and it computes the inner product in lines \ref{alg:ftg:condition}, \ref{alg:ftg:gram} and \ref{alg:ftg:alpha}.
In line \ref{alg:ftg:condition} only one computation of the inner product is made.
In line \ref{alg:ftg:gram}  algorithm makes $2k - 1$ evaluations of the inner product to add the last row and last column to the Gram matrix.
In line \ref{alg:ftg:alpha} algorithm makes another evaluation of the inner product $\inner{\cl{F}}{\cl{v_k}}$.

\subsection{Geometrical Interpretation of FTG} \label{sec:ftg-geom}

\begin{figure}[!tb]
    \centering
    \begin{subfigure}[t]{0.4\textwidth}
        \centering
        \includegraphics[width=\textwidth]{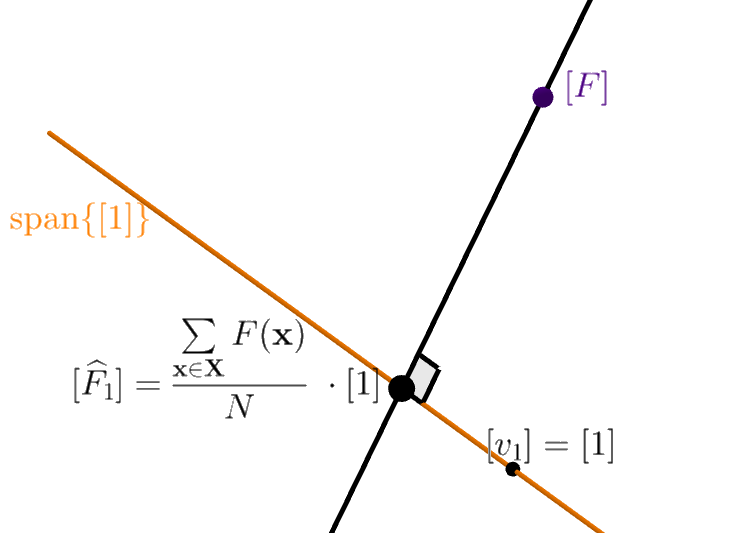}
        \caption{The first step of FTG for $k = 1$. The constant $1$ is added as the function $v_1$ and the best approximation of $\F$ with this constant is found as function $\Fest_1$.
        }
    \end{subfigure}
    \hfill
    \begin{subfigure}[t]{0.45\textwidth}
        \centering
        \includegraphics[width=\textwidth]{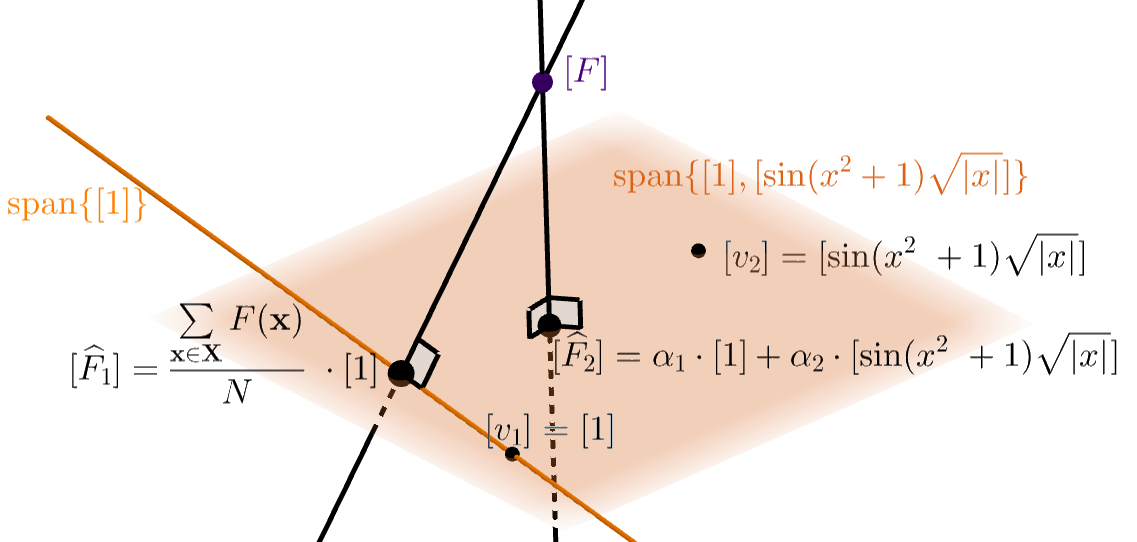}
        \caption{The second possible step of FTG for $k = 2$. We assume that a function $v_2 = \sin(x^2 + 1) \sqrt{|x|}$ is generated and the training data was such that $\cl{v_2}$ does not lie in the space spanned by $\cl{v_1}$.
        }
    \end{subfigure}
    \caption{Visualization of the first (a) and second (b) steps that FTG can make for a one-dimensional domain.}
    \label{fig:ftg-vis}
\end{figure}

In this section, we provide an example of FTG execution and give a visually intuitive scheme of what FTG algorithm does.

\begin{example}[First two steps of FTG on a one-dimensional domain]
FTG starts in line \ref{alg:ftg:constant-1} of Algorithm \ref{alg:ftg} with a single element, which is a constant 1. 
The class of equivalent functions that this constant represents is $\cl{1}$.
This case is shown in Figure \ref{fig:ftg-vis} (a), where the $\Span{\cl{1}}$ is shown with an orange line in the considered Hilbert space \F. 
The solution of LLSQ with only one function $v_1(\BF{x}) = 1$ gives the constant $\alpha$ such that $\norm{F - \alpha}$ is the smallest possible.
Substitution of this $v_1$ to Eq. \eqref{eq:prj_to_subspace} gives the following solution: $\alpha = {N}^{-1}{\sum_{\BF{y} \in \X}\F(\BF{y})}$.
The point $\alpha \cdot \cl{1}$ belongs to $\Span{\cl{1}}$ and it is closest point to $\cl{\F}$ in this subspace.
Moreover, the vector $\cl{\F} - \alpha \cdot \cl{1}$ is orthogonal to the subspace $\Span{\cl{1}}$ as shown in Figure \ref{fig:ftg-vis} (a).

After finding $\Fest_1$, FTG proceeds with $k = 2$ and enters a while loop in line \ref{alg:ftg:loop} to generate a composition of elementary functions.
Assume that this composition is $v_2(x) = \sin(x^2 + 1) \sqrt{|x|}$ and the data set is chosen in such a way, that $\cl{v_2} \notin \Span{\cl{v_1}}$.
This situation is shown in Figure \ref{fig:ftg-vis} (b), where element $\cl{v_2}$ is not located on the orange line.
In this case, elements $\cl{v_1}$ and $\cl{v_2}$ are linearly independent, and hence the subspace which they span increases.
This subspace $\Span{\cl{v_1}, \cl{v_2}}$ is shown as orange plane.
The condition in line \ref{alg:ftg:condition} of Algorithm \ref{alg:ftg} checks if vectors $\cl{\F} - \alpha \cdot \cl{v_1}$ and $\cl{v_2}$ are perpendicular.
Since they are not perpendicular, the algorithm leaves the while loop and proceeds with the computation of the Gram matrix.
The next best approximation of $\F$ is found in the plane shown in orange in Figure \ref{fig:ftg-vis} (b).
$\Fest_2 = \alpha_1 \cdot \cl{v_1} + \alpha_2 \cdot \cl{v_2}$, with constants $\alpha_1, \alpha_2$ obtained by application of Eq. \eqref{eq:prj_to_subspace} in line \ref{alg:ftg:alpha} of Algorithm \ref{alg:ftg}.

\label{ex:ftg-ex}
\end{example}

Example \ref{ex:ftg-ex} demonstrates how FTG handles randomly generated compositions of elementary operators.
FTG proceeds in this fashion, spanning greater subspaces, until the class of the target function \F is included in the spanned space.
The crucial point of the algorithm is the linear independence of the newly generated function with the previously generated ones.
Intuitive interpretation is that such a function adds new information about the target function.
Following this intuition, the predicate in line \ref{alg:ftg:condition} of Algorithm \ref{alg:ftg} checks if the generated composition adds any new information about the function.
When it is added, the known information is not lost, meaning the quality of the approximation is not reduced, which is guaranteed by Theorem \ref{theorem:4opt}.

\subsection{Implementation Details %
} \label{sec:impl}
\begin{algorithm}
\caption{Generate composition of elementary operators $\set{\mathfrak{U}, \mathfrak{B}, \mathfrak{C}, \Pi}$ given the minimal $l$ and maximal $u$ 
numbers of nested operators and a constant $p \in [0, 1]$. This is our adaptation of the classical algorithm for tree initialization, called ``ramped half-and-half'', see \cite{koza1992programming, ramped-half-and-half}.}
\begin{algorithmic}[1]
\Procedure{\textsc{generate-composition}}{$\mathfrak{U}, \mathfrak{B}, \mathfrak{C}, \Pi, p, l, u$}
    \State{$\widetilde{p} \sim U(\set{p, 1})$}
    \State{$\widetilde{u} \sim U\br{\Z_u \setminus \Z_{l-1}}$}
    \State{$d_1 \gets 0$; $i \gets 1$; $j \gets 2$}
    \State{$f_i \sim E\br{\widetilde{p}, d_i, l, \widetilde{u}}$}
    \While{$i < j$}\label{alg:cg:cur-time}
        \State{$d_j \gets d_i + 1$}
        \State{$d_{j+1} \gets d_i + 1$}
        \If {$ f_i \in \mathfrak{U}$}
            \State{$f_j \sim E\br{\widetilde{p}, d_i + 1, l, \widetilde{u}}$}
            \State{$f_i \gets f_i \circ f_j $} \label{alg:cg:unary-gen}
            \State{$j \gets j + 1$}
        \EndIf
        \If {$f_i \in \mathfrak{B}$}
            \State{$f_j \sim E\br{\widetilde{p}, d_i + 1, l, \widetilde{u}}$} 
            \State{$f_{j+1} \sim E\br{\widetilde{p}, d_i + 1, l, \widetilde{u}}$} \label{alg:cg:binary-gen-1}
            \State{$f_i \gets f_i(f_j, f_{j+1}) $} \label{alg:cg:binary-gen-2}
            \State{$j \gets j + 2$}
        \EndIf
        \State{$i \gets i + 1$} \label{alg:cg:cur-time}
    \EndWhile
    \Return{$f_1$ %
    }
\EndProcedure
\end{algorithmic}
\label{alg:generate-composition}
\end{algorithm}

In this section, we describe the practical aspects of the implementation of the proposed FTG algorithm.
We start by introducing the family of probability distributions parameterized by four real values $p,d,l,u$.
Every particular distribution in this family defines the distribution over elementary operators.
They are used to sample a particular elementary operator in our procedure to generate compositions.
Every sampled operator has associated depth $d$, which denotes the number of functions in which the operator is nested. 
The constants $l, u$ limit the minimal and maximal number of nested operators accordingly.
The constant $p$ defines the probability with which a unary or binary function will be sampled when this choice is possible.

\begin{equation}
    E\br{p, d, l, u} = 
    \begin{cases}
        U\br{\mathfrak{U} \cup \mathfrak{B}}, & \begin{small}\text{if } d < l \end{small}\\
        U\br{\mathfrak{U} \cup \mathfrak{B}}, & \begin{small}\text{with prob. } p \text { if } l \le d < u \end{small}\\
        U\br{U\br{\set{\Pi, \mathfrak{C}}}}, & \begin{small}\text{with prob. } 1 - p \text { if } l \le d < u \end{small} \\
        U\br{U\br{\set{\Pi, \mathfrak{C}}}}, & \begin{small}\text{if } d \ge u \end{small}
    \end{cases}
\end{equation}

Given this family of distribution $E$, we are ready to formulate the algorithm that generates a composition of functions. 
FTG utilizes our adaptation of ramped half-and-half initialization as a well-established tree-initialization method commonly used in GP to generate new compositions~\cite{koza1992programming, ramped-half-and-half}. 
We summarize this conventional methodology in Algorithm \ref{alg:generate-composition}.

Application of Algorithm \ref{alg:generate-composition} as function \textsc{generate-composition} is one of the possible ways to generate the composition of functions. 
We choose this implementation, because of its simplicity and unbiasedness between operators.

When a matrix is ill-conditioned, meaning it has a high condition number, the computation of its inverse is prone to numerical errors. %
However, in the area of approximation theory, it is known that arbitrary choice of linearly independent elements of Hilbert space will likely lead to a Gram matrix with a very high condition number \cite{holmes1991random, taylor1978condition}.
In our work, the functions, that specify elements of Hilbert space are produced randomly and independently from each other, so it is very likely that FTG struggles with such ill-conditioned Gram matrices.
In order to address this practical limitation, we compute the inverse using Singular Value Decomposition (SVD) and check if the inverse of the Gram matrix is close to its actual inverse. 
If this condition is not satisfied, then we do not include the generated $\cl{v_k}$ to the set of linearly independent elements and return to the line \ref{alg:ftg:condition} of Algorithm \ref{alg:ftg} to generate another $v_k$.
The check for closeness to the inverse is implemented as follows.
When an approximation $\widehat{\BF{G}^{-1}}$ of matrix $\BF{G}^{-1}$ is obtained, we consider $\BF{\widehat{I}} \coloneqq \BF{G}\cdot \widehat{\BF{G}^{-1}}$ and validate that every element of this product is close to the corresponding element of the identity matrix $\BF{I}$.
More precisely, for every $i,j$ we check that $|\widehat{I}_{i,j} - I_{i,j}| < \eps_1$.
In this paper, we use $\eps_1 = 10^{-4}$.

Such a procedure with a generation of $v_k$ adds additional computational complexity, but helps to avoid significant numerical errors with ill-conditioned matrices.
In our work, FTG was dealing with condition numbers up to $3.7 \cdot 10^{17}$ in average across all the considered conventional benchmarks and runs.

In our implementation, we used static parameters $p = 0.5, l = 1, u = 9$ for Algorithm \ref{alg:generate-composition}.
We assumed that the inner product defined in line \ref{alg:ftg:condition} of Algorithm \ref{alg:ftg} is zero when its absolute value is less than $\eps_2 = 10^{-3}$.
The bigger this constant is, the more difficult it is for the algorithm to generate the composition that satisfies the defined predicate.
On the other hand, the greater the constant, the more new element generated from the span of elements $v_1$, $v_2$, $\dots$, $v_{k-1}$, and thus the less ill-conditioned the Gram matrix is.
Therefore, the trade-off between $\eps_1$ and $\eps_2$ exists and must be identified. 
In our work, we selected those constants by trying multiple values.
We leave a more detailed investigation of this trade-off for future works.
The full code of the Algorithm \ref{alg:ftg}, Algorithm \ref{alg:generate-composition}, conventional GP algorithms that we considered, and all experiments we performed can be found online at~\cite{FTGcode}.

\section{Experiments}
\label{sec:evaluation}

\subsection{Experimental Setup}
\label{app:exp}

We performed experiments on classical one-dimensional symbolic regression problems. 
To evaluate the search performance for GP and our proposed algorithm, we measured the number evaluations of loss functionsons before the computational budget was exceeded. %
In addition to the mean values of the measurements, we calculated the standard deviation (SD) and the standard error of the mean (SEM). 
Binary tournament selection was used to select new parent individuals. The configuration for canonical GP was adopted from~\cite{10.1145/3583133.3596332}.  \\ 
We performed 100 independent  %
runs with different random seeds. We used the function defined in goal \ref{goal:minimizer} as the fitness %
function. 
When the difference of all absolute values becomes less than $\epsilon$, the algorithm is classified as converged. 
We considered the following values of $\epsilon$ from $10^{0}$ to $10^{-8}$ to evaluate different tolerance levels.
In our experiments we used a function set \[\mathfrak{B} = \{+, -, *, /, \sin, \cos, \ln\}\] and for each run we allowed a budget of $10^5$ fitness evaluations. Besides evaluating the conventional recombination-based GP, which we refer to as canonical GP, we also considered mutation-only GP that is used with a $(1+\lambda)$-EA and is referred to as $(1+\lambda)$-GP. 
Ramped half-and-half initialization has been used for all tested algorithms. The configuration of the respective GP algorithms is shown in Table~\ref{meta-eval-parameters} (in the Appendix). We divide our experiments into two parts. We evaluate the search performance for GP and FTG on conventional benchmarks that have been proposed for SR. 
  
\subsection{Conventional Benchmarks}

We selected nine well-known symbolic regression benchmarks from the work of McDermott et al.~\cite{DBLP:conf/gecco/McDermottWLMCVJKHJO12}. The objective functions and dataset configuration of the respective problems are shown in Table~\ref{problems_regression} (in the Appendix).

\subsection{Large-Scale Polynomial Benchmark}

We propose a type of benchmark for SR to address the following two points that are missing in our conventional SR benchmark.
\begin{enumerate}
    \item Finite training data set leaves a chance that the obtained expression with a small value of the loss function approximate value of $\F(\BF{x})$ well only for $\BF{x} \in \X$, but give high $f_{itness}$ value for points in $\domain \setminus \X$; \label{pt:overfit}
    \item GP applied to conventional benchmarking problems generates a function more complicated than the target function very quickly. 
    The span of such candidate solutions already includes the target function. This hinders observing the capabilities of algorithms to iteratively span subspaces that are getting closer to the target function \F. \label{pt:span}
\end{enumerate}

The proposed Large-Scale Polynomial (LSP) benchmark addresses Point \ref{pt:overfit} by considering the whole domain as the training data. 
This makes the computation of the loss function technically more difficult.
Point \ref{pt:span} is addressed by considering a target function as a polynomial of a high degree, for example, 100.
This entails problems for the computations of the loss values for candidate solutions, because of limitations in floating point precision on the computer.
Now we show how both mentioned difficulties, namely computation of the loss function over an infinite data set and possible numerical errors, are tackled in the proposed LSP benchmarks.

Given constants $a, b, k, c_0, c_1, \dotsc, c_k$, we define LSP benchmark, as SR instance, where $\mathfrak{B} =$$ \set{+, *}$, $ \mathfrak{U} = \varnothing$, $\Pi = \set{1}$,~$ \domain = [a, b]$,~$ \X = \domain$,~$ \F = \sum_{i=0}^kc_i x^i$. We focus on a particular case of our general setup when sets $\mathfrak{U,B}$ are the smallest that satisfy our assumptions.
I.e. we consider $\mathfrak{B} =$$ \set{+, *}$, $ \mathfrak{U} = \varnothing$, $\Pi = \set{1}$ which implies that the dimension of the problem is $n = 1$.
We denote such a one-dimensional domain as $\domain = [a, b]$.
Regardless of the set of constants $\mathfrak{C}$, all the compositions of elementary operators belong to the class of polynomials.
A polynomial of degree $k$ can always be written down in the form $\sum_{i=0}^k c_i x^i$, where $c_k \ne 0$.
We will denote this as \textit{normal form} of a polynomial.

Up until now, we considered only finite sets \X.
Now, we propose to consider infinite $\X$ which equals the whole domain where the target function \F is defined, i.e. $\X = \domain$.
In this case, we define $\H$ as space of square-integrable functions over the segment $(a,b)$, which we denote as $L_2(a, b)$.
Due to the fact that all the polynomials are square-integrable functions, such \H includes all compositions of the considered elementary operators.
$L_2(a, b)$ is a Hilbert space with inner product $\inner{f}{g} = \int_a^b{f(x)g(x)dx} \quad \forall f,g \in L_2(a,b)$.
It induces the following norm: $\norm{f} = \sqrt{\inner{f}{f}}$.

For this benchmark, we choose the target function \F as a polynomial. 
So the target function $\F \in L_2(a,b)$, hence the loss of SR, can be computed as $\mathcal{L}(\Fest) = \norm{\F - \Fest}^2 = \inner{\F - \Fest}{\F - \Fest}$.

It is clear, that being able to compute the value of the inner product automatically, is sufficient for automatic computation of loss value.
To implement this computation, we transform a tree to a polynomial in the form using a recursive algorithm.
Then we automatically do the subtraction of polynomials in normal form, if needed, and do the multiplication of polynomials in normal form.
After this, the resulting polynomial under the integral is obtained in normal form, and the computation of the inner product boils down to the computation of the following integral:
\begin{equation}\int\limits_a^b \sum\limits_{i=0}^kc_i x^i \mathrm{d}x = \sum\limits_{i=0}^k \dfrac{c_i}{i+1} (b^{i+1} - a^{i+1}) \label{eq:lsp-inner}\end{equation}

\subsection{Results}

\begin{figure*}[!tb]
     \centering
    \begin{tabular}{p{0.48\textwidth} p{0.48\textwidth}}
        \adjustbox{valign=t}{\includegraphics[width=0.4\textwidth,trim=0mm 0mm 40mm 7mm]{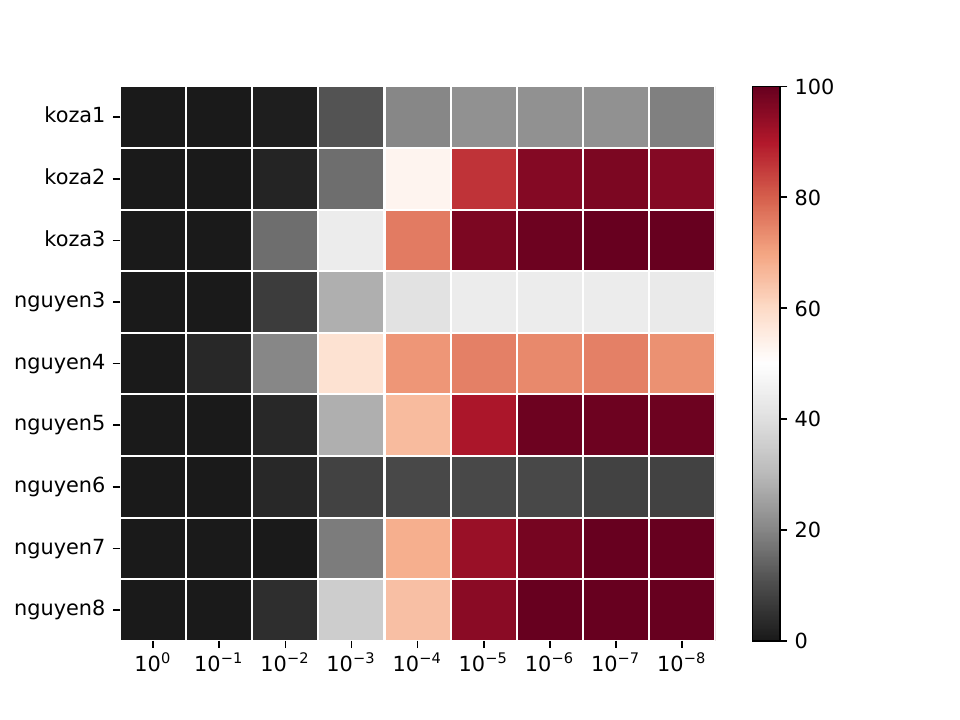}}
        &
        \adjustbox{valign=t}{\includegraphics[width=0.4\textwidth,trim=0mm 0mm 40mm 7mm]{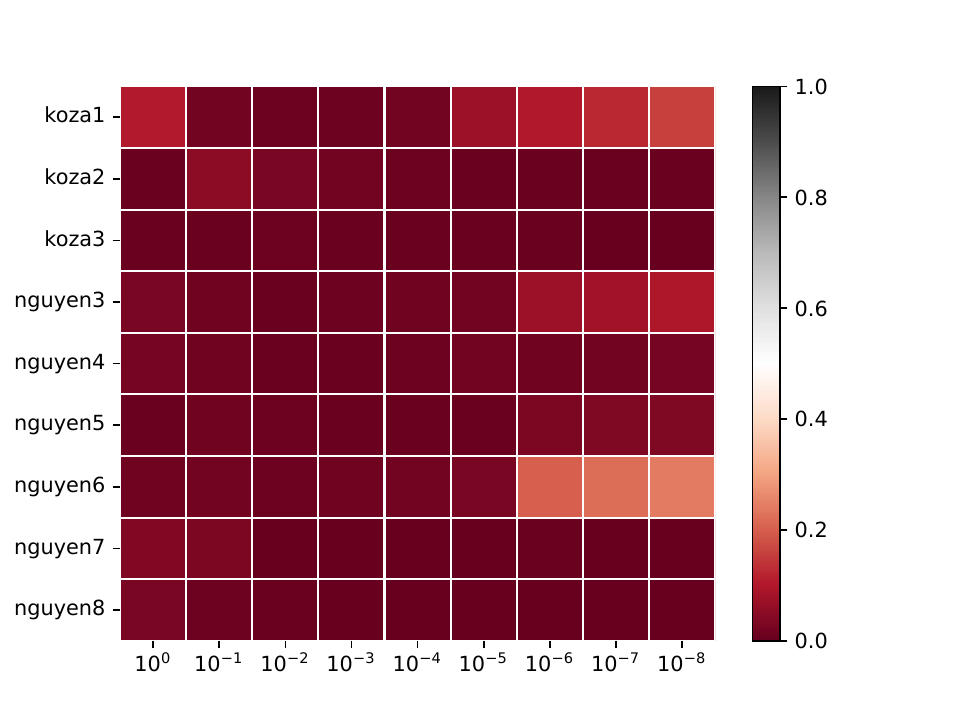}} \\
        \centering \textbf{(a) Difference of \textit{Success Rate}}
        &
        \centering \textbf{(b) Ratio of \textit{Median Function Evaluations}}
        \\
    \end{tabular}
    
    \caption{Comparison of success rate (a) and median of function evaluations to reach the given tolerance (b) of FTG with three standard GP algorithms: $(1+1)$-GP, $(1+\lambda)$-GP, Canonical-GP across a range of benchmark problems and tolerance levels. }
    \label{conventional_evaluation}
\end{figure*}

Figure~\ref{conventional_evaluation} shows the results of our experiments with conventional benchmarks. 
On each heatmap, the X-axis shows the tolerance value and the Y-axis shows the benchmark problem. 
In subfigure (a) every cell shows the difference between the success rate (in percent) of FTG and the maximal success rate that GP algorithms achieved.
In subfigure (b) every cell shows a ratio between median function evaluations of FTG and minimal value across medians of function evaluations that GP algorithms made to reach the given tolerance of the solution.
The more red the cell is, the better the performance of FTG relative to the considered GP algorithms. 
It is clearly visible that FTG performs considerably better and more robust for the tested problems when compared to conventional GP.  
The complete results are available in the Appendix in Table \ref{tab:rst2} and Table \ref{tab:rst1}.

\begin{figure}[]
    \centering
    \includegraphics[width=0.4\textwidth,trim=20mm 25mm 25mm 10mm]{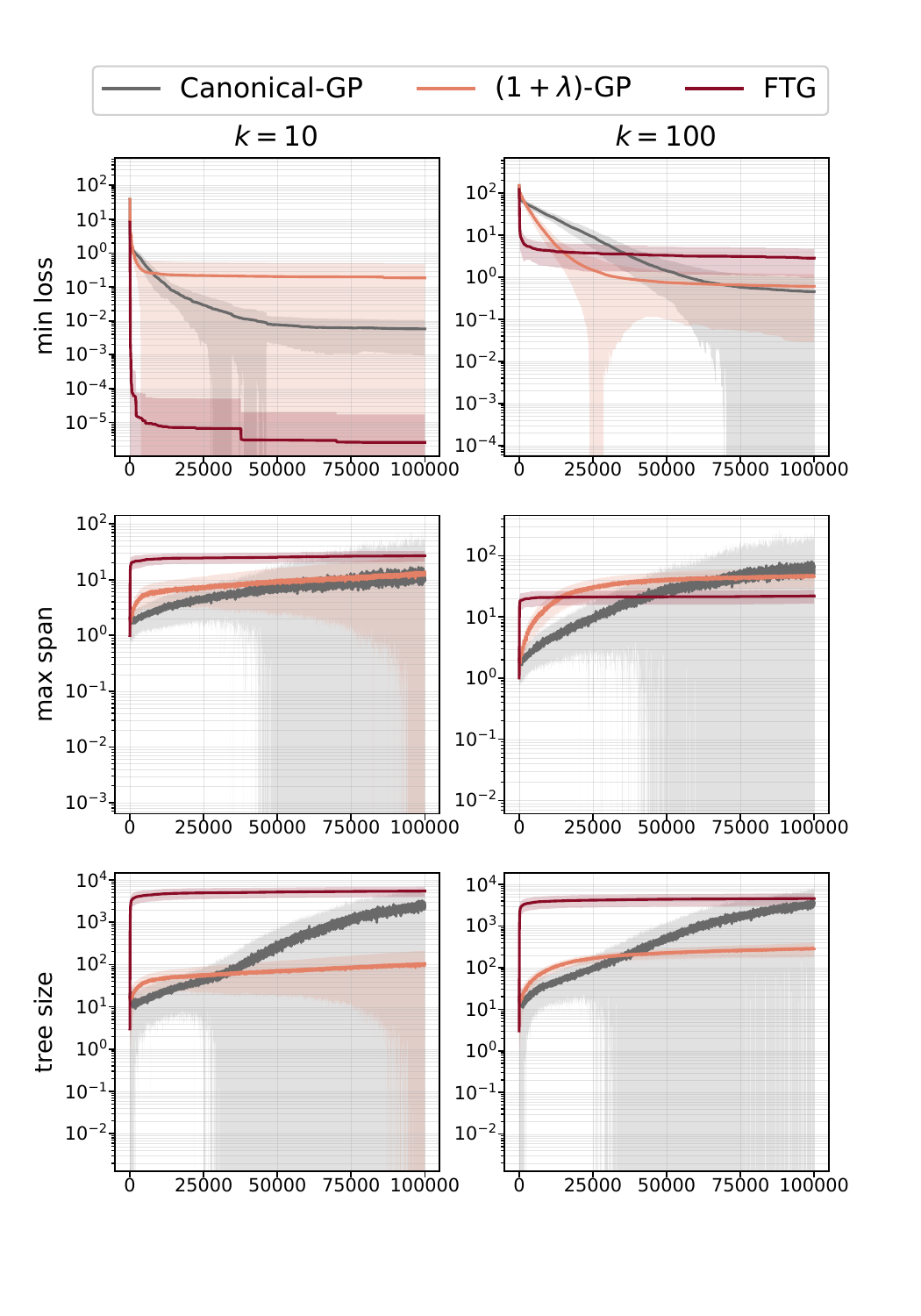}
    
    \caption{Performance Comparison of Canonical-GP
    (grey) and $(1+\lambda)$-GP
    (light red) and FTG (red) on the proposed Large-Scale Polynomial Benchmarks. 
    }
    \label{fig:large-scale-poly}
\end{figure}

Figure~\ref{fig:large-scale-poly} shows the results of the LSP evaluation. 
A column displays results for a different configuration of the benchmark, characterized by the polynomial degree $k$.
The target function is $\sum\nolimits_{i=0}^kx^i$. 
Rows represent the tracked quantity for each generation of the algorithm. 
Each chart depicts the averaged tracked quantity on the Y-axis and the generation number on the X-axis. 
The lines represent the mean value and the shaded areas represent the standard deviation. 
Finally, the last two rows with numbers show the means and standard deviations of the number of individuals that simultaneously achieved better fitness than the best-so-far solution and spanned a larger subspace. 
For every configuration of the experiment, we used the same random seeds and made $100$ independent runs to obtain statistically robust results.

In the further discussion, we say that a polynomial has the span of size $k$ when a polynomial in a normal form is $\sum_{i=0}^k c_i x^i$ such that $\forall i \in \Z_k: c_i \ne 0$.
The size of the span is measured automatically by transforming a tree to a polynomial in normal form and counting the number of terms with different degrees.

\section{Discussion}
\label{sec:discussion}
Figure \ref{conventional_evaluation} exhibits that FTG significantly outperforms GP in terms of the number of function evaluations needed to reach the desired precision of the solution. 
We highlight, that FTG reaches an almost absolute success rate even for very small tolerance levels. 
If the tolerance of the solution is infinitely large, then all the compositions of functions are accepted as sufficient approximation $\Fest$.
However, when the tolerance is reduced, the set of functions with sufficiently small loss values does not increase, and, for practically relevant elementary operators, it reduces.
Hence, for the minimal tolerance levels, the set of candidate solutions of sufficient quality is relatively small, which makes it harder for conventional GP to find any element from this set.
This results in a small success rate of GP observed in Figure \ref{conventional_evaluation}.
At the same time, FTG manages to overcome this challenge because it has an almost perfect success rate and spends a much smaller budget to find solutions for such tolerance. %
We rigorously prove in Theorem \ref{theorem:4opt} that, with the absence of numerical errors, FTG improves the quality of the approximation every time it manages to find a linearly independent function $v_k$.
GP search takes advantage of randomness and naturally evolves functions, which could be considered as a random walk in Hilbert space \H within the framework proposed in this work. %
Element $\cl{f}$ that is encountered during the random walk may be decomposed to the sum of elements if $f$ has addition as the upper functions in the composition.
Even if the top function in the composition is already not addition, element $\cl{f}$ belongs to the space $\Span{\cl{f}}$.
We can observe that when an element $\cl{f}$ is encountered during the random walk it may belong to a space that was already spanned earlier by element $\cl{g}$.
In this case, FTG filters out the function $f$, but GP may include it in the population.
Such suboptimal choice reduces the selective pressure towards more promising candidate solutions, which reduces the speed of convergence.

The first row of results on the LSP benchmark in Figure \ref{fig:large-scale-poly} exhibit that FTG quickly reduces loss value at the beginning, but then stagnates. FTG appears to struggle with the ill-conditionality of the Gram matrix, which is a hindrance for the accurate computation of its inverse, and in this way, numerical errors are caused.
According to our implementation of FTG as explained in Section \ref{sec:impl}, we loop until the inverse of the Gram matrix can be computed accurately, which causes stagnation of FTG when the Gram matrix becomes bigger.

In the first and second rows of Figure \ref{fig:large-scale-poly} we can see that GP spans greater subspaces and, at the same time, reduces loss value at a slower pace than FTG.
For the polynomial of degree 100, GP manages to eventually overcome FTG.
We propose the following explanation of the observed results:.
In general, mutation and recombination with certain non-zero probability change constants are hidden inside the candidate solution function $f$, while preserving the structure of this function.
Consider an individual as a function of the domain variable $\BF{x}$ and the vector of all constants $\Theta$, i.e., $f(\BF{x}, \Theta)$.
For changing $\Theta$, elements $\cl{f(\cdot, \Theta)}$ span some subspace $\mathcal{S}$, which, in the general case, can be neither convex nor complete in \H.
However, in $\mathcal{S}$, there might be many points such that $\cl{F}$ is closer to them than to the projection of $\F$ to the linear subspace $\Span{f(\cdot, \Theta_1)}$ for some fixed $\Theta_1$.
The mechanism that GP uses to generate new individuals allows it to obtain such points.

Surprisingly, we can observe that $(1+\lambda)$-GP, which uses only mutation to variate candidate solutions, suffers less from bloat, but converges faster than Canonical-GP, while having approximately the same loss and span of \H at the end of optimization.
It means that the mutation-only variation operator is beneficial for polynomials, but the study of how well it generalizes on other problems is left for future work.

LSP with a polynomial of degree 10 is used to see the performance of algorithms in the absence of overfitting.
The proposed FTG algorithm performs better than conventional GP on this instance of LSP but does not reach a loss value of zero. %
At the same time, in the last row of Figure \ref{fig:large-scale-poly} we see that FTG produces trees with a significantly greater number of nodes on both considered cases of polynomials.
This is due to the method that we used for the generation of compositions of functions in FTG.

In both considered cases of polynomials, Canonical-GP spans approximately the same number of dimensions that the target function spans.
Hence, GP found the approximate shape of the target polynomial but struggled to find the right constants.

Our analyses of GP on this benchmark demonstrate that GP excels in generating solutions that effectively span subspaces of the considered Hilbert space. 

\section{Conclusions and Future Work}
\label{sec:conclusion}
In this work, we considered SR from the perspective of functional analysis and proposed a novel algorithm that is able to navigate and optimize in Hilbert space. Our consideration of SR in Hilbert space allows us to achieve insight into the principles of conventional GP. 
To the best of our knowledge, our work represents a pioneering application of functional analysis in the SR domain, for which we proposed first theoretical and experimental results. 
Our proposed FTG algorithm demonstrates significant performance gains over considered GP algorithms on conventional benchmarks. 
FTG manages to find solutions of minimal tolerance, which shows that it can efficiently navigate in a large search space.
However, it is susceptible to numerical errors when the Gram matrix is inverted. We propose to address this issue by employing the Gram-Schmidt orthogonalization algorithm to transform linearly independent vectors into an orthonormal basis. 
The better performance of FTG on conventional benchmarks can be explained by the fact that FTG filters out a priory suboptimal functions, while GP does not. However, the working principles of GP allow it to be more beneficial for polynomials of high degree, which we show in the proposed LSP benchmarks.
In future work, we plan to hybridize GP and FTG to obtain an algorithm that can simultaneously substantially span the subspaces of the Hilbert space and effectively optimize the constants. 

\paragraph{\textbf{Acknowledgments}}
We are grateful to Dr. André Deutz, Leiden University, for the insightful discussions during the writing process and for his valuable detailed feedback on the paper. 
We also deeply appreciate the stimulating conversation with Prof. Dr. Günter Rudolph, TU Dortmund University, whose general view on our work greatly benefited our research.
This work is supported by the Austrian Science Fund (FWF – Der Wissenschaftsfonds) under the project (I 5315, `ML Methods for Feature Identification Global Optimization). The project was financially supported by ANR project HQI ANR-22-PNCQ-0002.

\bibliographystyle{ACM-Reference-Format}
\bibliography{mybibliography}


\begin{thebibliography}{46}


\ifx \showCODEN    \undefined \def \showCODEN     #1{\unskip}     \fi
\ifx \showDOI      \undefined \def \showDOI       #1{#1}\fi
\ifx \showISBNx    \undefined \def \showISBNx     #1{\unskip}     \fi
\ifx \showISBNxiii \undefined \def \showISBNxiii  #1{\unskip}     \fi
\ifx \showISSN     \undefined \def \showISSN      #1{\unskip}     \fi
\ifx \showLCCN     \undefined \def \showLCCN      #1{\unskip}     \fi
\ifx \shownote     \undefined \def \shownote      #1{#1}          \fi
\ifx \showarticletitle \undefined \def \showarticletitle #1{#1}   \fi
\ifx \showURL      \undefined \def \showURL       {\relax}        \fi
\providecommand\bibfield[2]{#2}
\providecommand\bibinfo[2]{#2}
\providecommand\natexlab[1]{#1}
\providecommand\showeprint[2][]{arXiv:#2}

\bibitem[Affenzeller et~al\mbox{.}(2014)]%
        {affenzeller2014gaining}
\bibfield{author}{\bibinfo{person}{Michael Affenzeller},
  \bibinfo{person}{Stephan~M Winkler}, \bibinfo{person}{Gabriel Kronberger},
  \bibinfo{person}{Michael Kommenda}, \bibinfo{person}{Bogdan Burlacu}, {and}
  \bibinfo{person}{Stefan Wagner}.} \bibinfo{year}{2014}\natexlab{}.
\newblock \showarticletitle{Gaining deeper insights in symbolic regression}.
\newblock \bibinfo{journal}{\emph{Genetic Programming Theory and Practice XI}}
  (\bibinfo{year}{2014}), \bibinfo{pages}{175--190}.
\newblock


\bibitem[Altenberg(1994)]%
        {kinnear:altenberg}
\bibfield{author}{\bibinfo{person}{Lee Altenberg}.}
  \bibinfo{year}{1994}\natexlab{}.
\newblock \showarticletitle{The Evolution of Evolvability in Genetic
  Programming}.
\newblock In \bibinfo{booktitle}{\emph{Advances in Genetic Programming}},
  \bibfield{editor}{\bibinfo{person}{Kenneth~E. {Kinnear, Jr.}}} (Ed.).
  \bibinfo{publisher}{MIT Press}, Chapter~3, \bibinfo{pages}{47--74}.
\newblock
\urldef\tempurl%
\url{https://doi.org/doi:10.7551/mitpress/1108.003.0009}
\showDOI{\tempurl}


\bibitem[Angeline(1997)]%
        {DBLP:conf/eps/Angeline97}
\bibfield{author}{\bibinfo{person}{Peter~J. Angeline}.}
  \bibinfo{year}{1997}\natexlab{}.
\newblock \showarticletitle{Comparing Subtree Crossover with Macromutation}. In
  \bibinfo{booktitle}{\emph{Evolutionary Programming VI, 6th International
  Conference, EP97, Indianapolis, Indiana, USA, April 13-16, 1997,
  Proceedings}} \emph{(\bibinfo{series}{Lecture Notes in Computer Science},
  Vol.~\bibinfo{volume}{1213})}, \bibfield{editor}{\bibinfo{person}{Peter~J.
  Angeline}, \bibinfo{person}{Robert~G. Reynolds}, \bibinfo{person}{John~R.
  McDonnell}, {and} \bibinfo{person}{Russell~C. Eberhart}} (Eds.).
  \bibinfo{publisher}{Springer}, \bibinfo{pages}{101--112}.
\newblock
\urldef\tempurl%
\url{https://doi.org/10.1007/BFB0014804}
\showDOI{\tempurl}


\bibitem[Antonov et~al\mbox{.}(2024)]%
        {FTGcode}
\bibfield{author}{\bibinfo{person}{Kirill Antonov}, \bibinfo{person}{Roman
  Kalkreuth}, \bibinfo{person}{Kaifeng Yang}, \bibinfo{person}{Thomas
  B{\"a}ck}, \bibinfo{person}{Niki van Stein}, {and} \bibinfo{person}{Anna~V
  Kononova}.} \bibinfo{year}{2024}\natexlab{}.
\newblock \bibinfo{title}{Source code of Fourier Tree Growing (FTG) and related
  experiments with Genetic Programming (GP)}.
\newblock
\newblock
\urldef\tempurl%
\url{https://anonymous.4open.science/r/fourier-tree-growing-46EB/README.md}
\showURL{%
\tempurl}


\bibitem[Cazenave(2013)]%
        {cazenave2013monte}
\bibfield{author}{\bibinfo{person}{Tristan Cazenave}.}
  \bibinfo{year}{2013}\natexlab{}.
\newblock \showarticletitle{Monte-carlo expression discovery}.
\newblock \bibinfo{journal}{\emph{International Journal on Artificial
  Intelligence Tools}} \bibinfo{volume}{22}, \bibinfo{number}{01}
  (\bibinfo{year}{2013}), \bibinfo{pages}{1250035}.
\newblock


\bibitem[Cozad and Sahinidis(2018)]%
        {cozad2018global}
\bibfield{author}{\bibinfo{person}{Alison Cozad} {and}
  \bibinfo{person}{Nikolaos~V Sahinidis}.} \bibinfo{year}{2018}\natexlab{}.
\newblock \showarticletitle{A global MINLP approach to symbolic regression}.
\newblock \bibinfo{journal}{\emph{Mathematical Programming}}
  \bibinfo{volume}{170} (\bibinfo{year}{2018}), \bibinfo{pages}{97--119}.
\newblock


\bibitem[Cramer(1985)]%
        {cramer1985representation}
\bibfield{author}{\bibinfo{person}{Nichael~Lynn Cramer}.}
  \bibinfo{year}{1985}\natexlab{}.
\newblock \showarticletitle{A representation for the Adaptive Generation of
  Simple Sequential Programs}. In \bibinfo{booktitle}{\emph{Proceedings of an
  International Conference on Genetic Algorithms and the Applications}},
  \bibfield{editor}{\bibinfo{person}{John~J. Grefenstette}} (Ed.).
  \bibinfo{address}{Carnegie-Mellon University, Pittsburgh, PA, USA},
  \bibinfo{pages}{183--187}.
\newblock
\urldef\tempurl%
\url{http://www.cs.ucl.ac.uk/staff/W.Langdon/ftp/papers/icga1985/icga85_cramer.pdf}
\showURL{%
\tempurl}


\bibitem[de~Fran{\c{c}}a(2018)]%
        {de2018greedy}
\bibfield{author}{\bibinfo{person}{Fabr{\'\i}cio~Olivetti de Fran{\c{c}}a}.}
  \bibinfo{year}{2018}\natexlab{}.
\newblock \showarticletitle{A greedy search tree heuristic for symbolic
  regression}.
\newblock \bibinfo{journal}{\emph{Information Sciences}}  \bibinfo{volume}{442}
  (\bibinfo{year}{2018}), \bibinfo{pages}{18--32}.
\newblock


\bibitem[Falkenhainer and Michalski(1986)]%
        {falkenhainer1986integrating}
\bibfield{author}{\bibinfo{person}{Brian~C Falkenhainer} {and}
  \bibinfo{person}{Ryszard~S Michalski}.} \bibinfo{year}{1986}\natexlab{}.
\newblock \showarticletitle{Integrating quantitative and qualitative discovery:
  the ABACUS system}.
\newblock \bibinfo{journal}{\emph{Machine Learning}}  \bibinfo{volume}{1}
  (\bibinfo{year}{1986}), \bibinfo{pages}{367--401}.
\newblock


\bibitem[Forsyth(1981)]%
        {doi:10.1108/eb005587}
\bibfield{author}{\bibinfo{person}{Richard Forsyth}.}
  \bibinfo{year}{1981}\natexlab{}.
\newblock \showarticletitle{{BEAGLE} A {Darwinian} Approach to Pattern
  Recognition}.
\newblock \bibinfo{journal}{\emph{Kybernetes}} \bibinfo{volume}{10},
  \bibinfo{number}{3} (\bibinfo{year}{1981}), \bibinfo{pages}{159--166}.
\newblock
\showISSN{0368-492X}
\urldef\tempurl%
\url{https://doi.org/doi:10.1108/eb005587}
\showDOI{\tempurl}


\bibitem[Galv\'{a}n-L\'{o}pez et~al\mbox{.}(2010)]%
        {10.1145/1830483.1830646}
\bibfield{author}{\bibinfo{person}{Edgar Galv\'{a}n-L\'{o}pez},
  \bibinfo{person}{James McDermott}, \bibinfo{person}{Michael O'Neill}, {and}
  \bibinfo{person}{Anthony Brabazon}.} \bibinfo{year}{2010}\natexlab{}.
\newblock \showarticletitle{Towards an Understanding of Locality in Genetic
  Programming}. In \bibinfo{booktitle}{\emph{Proceedings of the 12th Annual
  Conference on Genetic and Evolutionary Computation}} (Portland, Oregon, USA)
  \emph{(\bibinfo{series}{GECCO '10})}. \bibinfo{publisher}{Association for
  Computing Machinery}, \bibinfo{address}{New York, NY, USA},
  \bibinfo{pages}{901–908}.
\newblock
\showISBNx{9781450300728}
\urldef\tempurl%
\url{https://doi.org/10.1145/1830483.1830646}
\showDOI{\tempurl}


\bibitem[Gerwin(1974)]%
        {gerwin1974information}
\bibfield{author}{\bibinfo{person}{Donald Gerwin}.}
  \bibinfo{year}{1974}\natexlab{}.
\newblock \showarticletitle{Information processing, data inferences, and
  scientific generalization}.
\newblock \bibinfo{journal}{\emph{Behavioral Science}} \bibinfo{volume}{19},
  \bibinfo{number}{5} (\bibinfo{year}{1974}), \bibinfo{pages}{314--325}.
\newblock


\bibitem[Hicklin(1986)]%
        {Hicklin}
\bibfield{author}{\bibinfo{person}{Joseph Hicklin}.}
  \bibinfo{year}{1986}\natexlab{}.
\newblock \emph{\bibinfo{title}{Application of the Genetic Algorithm to
  Automatic Program Generation}}.
\newblock \bibinfo{thesistype}{Master's\ thesis}. \bibinfo{school}{University
  of Idaho}.
\newblock


\bibitem[Holmes(1991)]%
        {holmes1991random}
\bibfield{author}{\bibinfo{person}{Richard~B Holmes}.}
  \bibinfo{year}{1991}\natexlab{}.
\newblock \showarticletitle{On random correlation matrices}.
\newblock \bibinfo{journal}{\emph{SIAM journal on matrix analysis and
  applications}} \bibinfo{volume}{12}, \bibinfo{number}{2}
  (\bibinfo{year}{1991}), \bibinfo{pages}{239--272}.
\newblock


\bibitem[Hu and Banzhaf(2016)]%
        {DBLP:conf/gptp/HuB16}
\bibfield{author}{\bibinfo{person}{Ting Hu} {and} \bibinfo{person}{Wolfgang
  Banzhaf}.} \bibinfo{year}{2016}\natexlab{}.
\newblock \showarticletitle{Neutrality, Robustness, and Evolvability in Genetic
  Programming}. In \bibinfo{booktitle}{\emph{Genetic Programming Theory and
  Practice XIV, {[GPTP} 2016, University of Michigan, Ann Arbor, USA, May
  19-21, 2016]}} \emph{(\bibinfo{series}{Genetic and Evolutionary
  Computation})}, \bibfield{editor}{\bibinfo{person}{Rick~L. Riolo},
  \bibinfo{person}{Bill Worzel}, \bibinfo{person}{Brian Goldman}, {and}
  \bibinfo{person}{Bill Tozier}} (Eds.). \bibinfo{publisher}{Springer},
  \bibinfo{pages}{101--117}.
\newblock
\urldef\tempurl%
\url{https://doi.org/10.1007/978-3-319-97088-2\_7}
\showDOI{\tempurl}


\bibitem[Jackson(2010)]%
        {jackson2010identification}
\bibfield{author}{\bibinfo{person}{David Jackson}.}
  \bibinfo{year}{2010}\natexlab{}.
\newblock \showarticletitle{The identification and exploitation of dormancy in
  genetic programming}.
\newblock \bibinfo{journal}{\emph{Genetic Programming and Evolvable Machines}}
  \bibinfo{volume}{11} (\bibinfo{year}{2010}), \bibinfo{pages}{89--121}.
\newblock


\bibitem[Kammerer et~al\mbox{.}(2020)]%
        {kammerer2020symbolic}
\bibfield{author}{\bibinfo{person}{Lukas Kammerer}, \bibinfo{person}{Gabriel
  Kronberger}, \bibinfo{person}{Bogdan Burlacu}, \bibinfo{person}{Stephan~M
  Winkler}, \bibinfo{person}{Michael Kommenda}, {and} \bibinfo{person}{Michael
  Affenzeller}.} \bibinfo{year}{2020}\natexlab{}.
\newblock \showarticletitle{Symbolic regression by exhaustive search: Reducing
  the search space using syntactical constraints and efficient semantic
  structure deduplication}.
\newblock \bibinfo{journal}{\emph{Genetic programming theory and practice
  XVII}} (\bibinfo{year}{2020}), \bibinfo{pages}{79--99}.
\newblock


\bibitem[Kantorovich and Akilov(2016)]%
        {kantorovich2016functional}
\bibfield{author}{\bibinfo{person}{Leonid~Vital'evich Kantorovich} {and}
  \bibinfo{person}{Gleb~Pavlovich Akilov}.} \bibinfo{year}{2016}\natexlab{}.
\newblock \bibinfo{booktitle}{\emph{Functional analysis}}.
\newblock \bibinfo{publisher}{Elsevier}.
\newblock


\bibitem[Kotanchek et~al\mbox{.}(2013)]%
        {kotanchek2013symbolic}
\bibfield{author}{\bibinfo{person}{Mark~E Kotanchek},
  \bibinfo{person}{Ekaterina Vladislavleva}, {and} \bibinfo{person}{Guido
  Smits}.} \bibinfo{year}{2013}\natexlab{}.
\newblock \showarticletitle{Symbolic regression is not enough: it takes a
  village to raise a model}.
\newblock \bibinfo{journal}{\emph{Genetic Programming Theory and Practice X}}
  (\bibinfo{year}{2013}), \bibinfo{pages}{187--203}.
\newblock


\bibitem[Koza(1990)]%
        {koza1990genetic}
\bibfield{author}{\bibinfo{person}{J. Koza}.} \bibinfo{year}{1990}\natexlab{}.
\newblock \bibinfo{booktitle}{\emph{Genetic {P}rogramming: {A} paradigm for
  genetically breeding populations of computer programs to solve problems}}.
\newblock \bibinfo{type}{Technical Report} {STAN}-{CS}-90-1314.
  \bibinfo{institution}{Dept. of Computer Science, Stanford University}.
\newblock


\bibitem[Koza(1992a)]%
        {koza1992programming}
\bibfield{author}{\bibinfo{person}{JRGP Koza}.}
  \bibinfo{year}{1992}\natexlab{a}.
\newblock \showarticletitle{On the programming of computers by means of natural
  selection}.
\newblock \bibinfo{journal}{\emph{Genetic programming}} (\bibinfo{year}{1992}).
\newblock


\bibitem[Koza(1992b)]%
        {Koza:1992:GPP:138936}
\bibfield{author}{\bibinfo{person}{John~R. Koza}.}
  \bibinfo{year}{1992}\natexlab{b}.
\newblock \bibinfo{booktitle}{\emph{Genetic Programming: On the Programming of
  Computers by Means of Natural Selection}}.
\newblock \bibinfo{publisher}{MIT Press}, \bibinfo{address}{Cambridge, MA,
  USA}.
\newblock
\showISBNx{0-262-11170-5}
\urldef\tempurl%
\url{http://mitpress.mit.edu/books/genetic-programming}
\showURL{%
\tempurl}


\bibitem[Koza(1994)]%
        {koza1994genetic}
\bibfield{author}{\bibinfo{person}{John~R. Koza}.}
  \bibinfo{year}{1994}\natexlab{}.
\newblock \bibinfo{booktitle}{\emph{Genetic Programming II: Automatic Discovery
  of Reusable Programs}}.
\newblock \bibinfo{publisher}{MIT Press}, \bibinfo{address}{Cambridge
  Massachusetts}.
\newblock
\showISBNx{0-262-11189-6}
\urldef\tempurl%
\url{http://www.genetic-programming.org/gpbook2toc.html}
\showURL{%
\tempurl}


\bibitem[Kronberger et~al\mbox{.}(2018)]%
        {kronberger2018predicting}
\bibfield{author}{\bibinfo{person}{Gabriel Kronberger},
  \bibinfo{person}{Michael Kommenda}, \bibinfo{person}{Andreas Promberger},
  {and} \bibinfo{person}{Falk Nickel}.} \bibinfo{year}{2018}\natexlab{}.
\newblock \showarticletitle{Predicting friction system performance with
  symbolic regression and genetic programming with factor variables}. In
  \bibinfo{booktitle}{\emph{Proceedings of the Genetic and Evolutionary
  Computation Conference}}. \bibinfo{pages}{1278--1285}.
\newblock


\bibitem[Langley(1981)]%
        {langley1981data}
\bibfield{author}{\bibinfo{person}{Pat Langley}.}
  \bibinfo{year}{1981}\natexlab{}.
\newblock \showarticletitle{Data-driven discovery of physical laws}.
\newblock \bibinfo{journal}{\emph{Cognitive Science}} \bibinfo{volume}{5},
  \bibinfo{number}{1} (\bibinfo{year}{1981}), \bibinfo{pages}{31--54}.
\newblock


\bibitem[Lissovoi and Oliveto(2018)]%
        {DBLP:conf/aaai/LissovoiO18}
\bibfield{author}{\bibinfo{person}{Andrei Lissovoi} {and}
  \bibinfo{person}{Pietro~S. Oliveto}.} \bibinfo{year}{2018}\natexlab{}.
\newblock \showarticletitle{On the Time and Space Complexity of Genetic
  Programming for Evolving Boolean Conjunctions}. In
  \bibinfo{booktitle}{\emph{Proceedings of the Thirty-Second {AAAI} Conference
  on Artificial Intelligence, (AAAI-18), the 30th innovative Applications of
  Artificial Intelligence (IAAI-18), and the 8th {AAAI} Symposium on
  Educational Advances in Artificial Intelligence (EAAI-18), New Orleans,
  Louisiana, USA, February 2-7, 2018}},
  \bibfield{editor}{\bibinfo{person}{Sheila~A. McIlraith} {and}
  \bibinfo{person}{Kilian~Q. Weinberger}} (Eds.). \bibinfo{publisher}{{AAAI}
  Press}, \bibinfo{pages}{1363--1370}.
\newblock
\urldef\tempurl%
\url{https://doi.org/10.1609/AAAI.V32I1.11517}
\showDOI{\tempurl}


\bibitem[Lissovoi and Oliveto(2020)]%
        {DBLP:series/ncs/LissovoiO20}
\bibfield{author}{\bibinfo{person}{Andrei Lissovoi} {and}
  \bibinfo{person}{Pietro~S. Oliveto}.} \bibinfo{year}{2020}\natexlab{}.
\newblock \showarticletitle{Computational Complexity Analysis of Genetic
  Programming}.
\newblock In \bibinfo{booktitle}{\emph{Theory of Evolutionary Computation -
  Recent Developments in Discrete Optimization}},
  \bibfield{editor}{\bibinfo{person}{Benjamin Doerr} {and}
  \bibinfo{person}{Frank Neumann}} (Eds.). \bibinfo{publisher}{Springer},
  \bibinfo{pages}{475--518}.
\newblock
\urldef\tempurl%
\url{https://doi.org/10.1007/978-3-030-29414-4\_11}
\showDOI{\tempurl}


\bibitem[Luke and Panait(2001)]%
        {ramped-half-and-half}
\bibfield{author}{\bibinfo{person}{Sean Luke} {and} \bibinfo{person}{Liviu
  Panait}.} \bibinfo{year}{2001}\natexlab{}.
\newblock \showarticletitle{A Survey and Comparison of Tree Generation
  Algorithms}.
\newblock \bibinfo{journal}{\emph{Proceedings of the Genetic and Evolutionary
  Computation Conference (GECCO-2001)}} (\bibinfo{date}{06}
  \bibinfo{year}{2001}).
\newblock


\bibitem[Mambrini and Oliveto(2016)]%
        {DBLP:conf/eurogp/MambriniO16}
\bibfield{author}{\bibinfo{person}{Andrea Mambrini} {and}
  \bibinfo{person}{Pietro~S. Oliveto}.} \bibinfo{year}{2016}\natexlab{}.
\newblock \showarticletitle{On the Analysis of Simple Genetic Programming for
  Evolving Boolean Functions}. In \bibinfo{booktitle}{\emph{Genetic Programming
  - 19th European Conference, EuroGP 2016, Porto, Portugal, March 30 - April 1,
  2016, Proceedings}} \emph{(\bibinfo{series}{Lecture Notes in Computer
  Science}, Vol.~\bibinfo{volume}{9594})},
  \bibfield{editor}{\bibinfo{person}{Malcolm~I. Heywood},
  \bibinfo{person}{James McDermott}, \bibinfo{person}{Mauro Castelli},
  \bibinfo{person}{Ernesto Costa}, {and} \bibinfo{person}{Kevin Sim}} (Eds.).
  \bibinfo{publisher}{Springer}, \bibinfo{pages}{99--114}.
\newblock
\urldef\tempurl%
\url{https://doi.org/10.1007/978-3-319-30668-1\_7}
\showDOI{\tempurl}


\bibitem[McDermott et~al\mbox{.}(2012)]%
        {DBLP:conf/gecco/McDermottWLMCVJKHJO12}
\bibfield{author}{\bibinfo{person}{James McDermott},
  \bibinfo{person}{David~Robert White}, \bibinfo{person}{Sean Luke},
  \bibinfo{person}{Luca Manzoni}, \bibinfo{person}{Mauro Castelli},
  \bibinfo{person}{Leonardo Vanneschi}, \bibinfo{person}{Wojciech Jaskowski},
  \bibinfo{person}{Krzysztof Krawiec}, \bibinfo{person}{Robin Harper},
  \bibinfo{person}{Kenneth A.~De Jong}, {and} \bibinfo{person}{Una{-}May
  O'Reilly}.} \bibinfo{year}{2012}\natexlab{}.
\newblock \showarticletitle{Genetic programming needs better benchmarks}. In
  \bibinfo{booktitle}{\emph{Genetic and Evolutionary Computation Conference,
  {GECCO} '12, Philadelphia, PA, USA, July 7-11, 2012}},
  \bibfield{editor}{\bibinfo{person}{Terence Soule} {and}
  \bibinfo{person}{Jason~H. Moore}} (Eds.). \bibinfo{publisher}{{ACM}},
  \bibinfo{pages}{791--798}.
\newblock
\urldef\tempurl%
\url{https://doi.org/10.1145/2330163.2330273}
\showDOI{\tempurl}


\bibitem[Miller(1999)]%
        {miller:1999:ACGP}
\bibfield{author}{\bibinfo{person}{Julian~F. Miller}.}
  \bibinfo{year}{1999}\natexlab{}.
\newblock \showarticletitle{An empirical study of the efficiency of learning
  {boolean} functions using a Cartesian Genetic Programming approach}. In
  \bibinfo{booktitle}{\emph{Proceedings of the Genetic and Evolutionary
  Computation Conference}}, \bibfield{editor}{\bibinfo{person}{Wolfgang
  Banzhaf}, \bibinfo{person}{Jason Daida}, \bibinfo{person}{Agoston~E. Eiben},
  \bibinfo{person}{Max~H. Garzon}, \bibinfo{person}{Vasant Honavar},
  \bibinfo{person}{Mark Jakiela}, {and} \bibinfo{person}{Robert~E. Smith}}
  (Eds.), Vol.~\bibinfo{volume}{2}. \bibinfo{publisher}{Morgan Kaufmann},
  \bibinfo{address}{Orlando, Florida, USA}, \bibinfo{pages}{1135--1142}.
\newblock
\showISBNx{1-55860-611-4}
\urldef\tempurl%
\url{http://citeseer.ist.psu.edu/153431.html}
\showURL{%
\tempurl}


\bibitem[Moraglio et~al\mbox{.}(2012)]%
        {DBLP:conf/ppsn/MoraglioKJ12}
\bibfield{author}{\bibinfo{person}{Alberto Moraglio},
  \bibinfo{person}{Krzysztof Krawiec}, {and} \bibinfo{person}{Colin~G.
  Johnson}.} \bibinfo{year}{2012}\natexlab{}.
\newblock \showarticletitle{Geometric Semantic Genetic Programming}. In
  \bibinfo{booktitle}{\emph{Parallel Problem Solving from Nature - {PPSN} {XII}
  - 12th International Conference, Taormina, Italy, September 1-5, 2012,
  Proceedings, Part {I}}} \emph{(\bibinfo{series}{Lecture Notes in Computer
  Science}, Vol.~\bibinfo{volume}{7491})},
  \bibfield{editor}{\bibinfo{person}{Carlos A.~Coello Coello},
  \bibinfo{person}{Vincenzo Cutello}, \bibinfo{person}{Kalyanmoy Deb},
  \bibinfo{person}{Stephanie Forrest}, \bibinfo{person}{Giuseppe Nicosia},
  {and} \bibinfo{person}{Mario Pavone}} (Eds.). \bibinfo{publisher}{Springer},
  \bibinfo{pages}{21--31}.
\newblock
\urldef\tempurl%
\url{https://doi.org/10.1007/978-3-642-32937-1\_3}
\showDOI{\tempurl}


\bibitem[Openshaw and Turton(1994)]%
        {Openshaw94buildingnew}
\bibfield{author}{\bibinfo{person}{S. Openshaw} {and} \bibinfo{person}{I.
  Turton}.} \bibinfo{year}{1994}\natexlab{}.
\newblock \showarticletitle{Building new spatial interaction models using
  genetic programming}. In \bibinfo{booktitle}{\emph{Evolutionary Computing,
  Lecture Notes in Computer Science}}. \bibinfo{publisher}{Springer-Verlag},
  \bibinfo{pages}{11--13}.
\newblock


\bibitem[Perkis(1994)]%
        {ieee94:perkis}
\bibfield{author}{\bibinfo{person}{Tim Perkis}.}
  \bibinfo{year}{1994}\natexlab{}.
\newblock \showarticletitle{Stack-Based Genetic Programming}. In
  \bibinfo{booktitle}{\emph{Proceedings of the 1994 IEEE World Congress on
  Computational Intelligence}}, Vol.~\bibinfo{volume}{1}.
  \bibinfo{publisher}{IEEE Press}, \bibinfo{address}{Orlando, Florida, USA},
  \bibinfo{pages}{148--153}.
\newblock
\urldef\tempurl%
\url{https://doi.org/doi:10.1109/ICEC.1994.350025}
\showDOI{\tempurl}


\bibitem[Poli(1996)]%
        {Poli1996}
\bibfield{author}{\bibinfo{person}{Riccardo Poli}.}
  \bibinfo{year}{1996}\natexlab{}.
\newblock \bibinfo{booktitle}{\emph{Parallel Distributed Genetic Programming}}.
\newblock \bibinfo{type}{Technical Report} CSRP-96-15.
  \bibinfo{institution}{School of Computer Science},
  \bibinfo{address}{University of Birmingham, B15 2TT, UK}.
\newblock
\urldef\tempurl%
\url{ftp://ftp.cs.bham.ac.uk/pub/tech-reports/1996/CSRP-96-15.ps.gz}
\showURL{%
\tempurl}


\bibitem[Reiter et~al\mbox{.}(2023)]%
        {10.1145/3583133.3596332}
\bibfield{author}{\bibinfo{person}{Johannes Reiter}, \bibinfo{person}{Dirk
  Schweim}, {and} \bibinfo{person}{David Wittenberg}.}
  \bibinfo{year}{2023}\natexlab{}.
\newblock \showarticletitle{Pretraining Reduces Runtime in Denoising
  Autoencoder Genetic Programming by an Order of Magnitude}. In
  \bibinfo{booktitle}{\emph{Proceedings of the Companion Conference on Genetic
  and Evolutionary Computation}} (Lisbon, Portugal)
  \emph{(\bibinfo{series}{GECCO '23 Companion})}.
  \bibinfo{publisher}{Association for Computing Machinery},
  \bibinfo{address}{New York, NY, USA}, \bibinfo{pages}{2382–2385}.
\newblock
\showISBNx{9798400701207}
\urldef\tempurl%
\url{https://doi.org/10.1145/3583133.3596332}
\showDOI{\tempurl}


\bibitem[Ryan et~al\mbox{.}(1998)]%
        {ryan:1998:geepal}
\bibfield{author}{\bibinfo{person}{Conor Ryan}, \bibinfo{person}{J.~J.
  Collins}, {and} \bibinfo{person}{Michael O'Neill}.}
  \bibinfo{year}{1998}\natexlab{}.
\newblock \showarticletitle{Grammatical Evolution: Evolving Programs for an
  Arbitrary Language}. In \bibinfo{booktitle}{\emph{Proceedings of the First
  European Workshop on Genetic Programming}} \emph{(\bibinfo{series}{LNCS},
  Vol.~\bibinfo{volume}{1391})}, \bibfield{editor}{\bibinfo{person}{Wolfgang
  Banzhaf}, \bibinfo{person}{Riccardo Poli}, \bibinfo{person}{Marc Schoenauer},
  {and} \bibinfo{person}{Terence~C. Fogarty}} (Eds.).
  \bibinfo{publisher}{Springer-Verlag}, \bibinfo{address}{Paris},
  \bibinfo{pages}{83--96}.
\newblock
\showISBNx{3-540-64360-5}
\urldef\tempurl%
\url{https://doi.org/doi:10.1007/BFb0055930}
\showDOI{\tempurl}


\bibitem[Soule et~al\mbox{.}(1996)]%
        {10.5555/1595536.1595563}
\bibfield{author}{\bibinfo{person}{Terence Soule}, \bibinfo{person}{James~A.
  Foster}, {and} \bibinfo{person}{John Dickinson}.}
  \bibinfo{year}{1996}\natexlab{}.
\newblock \showarticletitle{Code Growth in Genetic Programming}. In
  \bibinfo{booktitle}{\emph{Proceedings of the 1st Annual Conference on Genetic
  Programming}} (Stanford, California). \bibinfo{publisher}{MIT Press},
  \bibinfo{address}{Cambridge, MA, USA}, \bibinfo{pages}{215–223}.
\newblock
\showISBNx{0262611279}


\bibitem[Sun et~al\mbox{.}(2022)]%
        {sun2022symbolic}
\bibfield{author}{\bibinfo{person}{Fangzheng Sun}, \bibinfo{person}{Yang Liu},
  \bibinfo{person}{Jian-Xun Wang}, {and} \bibinfo{person}{Hao Sun}.}
  \bibinfo{year}{2022}\natexlab{}.
\newblock \showarticletitle{Symbolic physics learner: Discovering governing
  equations via monte carlo tree search}.
\newblock \bibinfo{journal}{\emph{arXiv preprint arXiv:2205.13134}}
  (\bibinfo{year}{2022}).
\newblock


\bibitem[Taylor(1978)]%
        {taylor1978condition}
\bibfield{author}{\bibinfo{person}{James~M Taylor}.}
  \bibinfo{year}{1978}\natexlab{}.
\newblock \showarticletitle{The condition of Gram matrices and related
  problems}.
\newblock \bibinfo{journal}{\emph{Proceedings of the Royal Society of Edinburgh
  Section A: Mathematics}} \bibinfo{volume}{80}, \bibinfo{number}{1-2}
  (\bibinfo{year}{1978}), \bibinfo{pages}{45--56}.
\newblock


\bibitem[Vanneschi et~al\mbox{.}(2004)]%
        {vanneschi:fca:gecco2004}
\bibfield{author}{\bibinfo{person}{Leonardo Vanneschi}, \bibinfo{person}{Manuel
  Clergue}, \bibinfo{person}{Philippe Collard}, \bibinfo{person}{Marco
  Tomassini}, {and} \bibinfo{person}{S\'ebastien V\'erel}.}
  \bibinfo{year}{2004}\natexlab{}.
\newblock \showarticletitle{Fitness Clouds and Problem Hardness in Genetic
  Programming}. In \bibinfo{booktitle}{\emph{Genetic and Evolutionary
  Computation -- GECCO-2004, Part II}} \emph{(\bibinfo{series}{Lecture Notes in
  Computer Science}, Vol.~\bibinfo{volume}{3103})},
  \bibfield{editor}{\bibinfo{person}{Kalyanmoy Deb}, \bibinfo{person}{Riccardo
  Poli}, \bibinfo{person}{Wolfgang Banzhaf}, \bibinfo{person}{Hans-Georg
  Beyer}, \bibinfo{person}{Edmund Burke}, \bibinfo{person}{Paul Darwen},
  \bibinfo{person}{Dipankar Dasgupta}, \bibinfo{person}{Dario Floreano},
  \bibinfo{person}{James Foster}, \bibinfo{person}{Mark Harman},
  \bibinfo{person}{Owen Holland}, \bibinfo{person}{Pier~Luca Lanzi},
  \bibinfo{person}{Lee Spector}, \bibinfo{person}{Andrea Tettamanzi},
  \bibinfo{person}{Dirk Thierens}, {and} \bibinfo{person}{Andy Tyrrell}}
  (Eds.). \bibinfo{publisher}{Springer-Verlag}, \bibinfo{address}{Seattle, WA,
  USA}, \bibinfo{pages}{690--701}.
\newblock
\showISBNx{3-540-22343-6}
\showISSN{0302-9743}
\urldef\tempurl%
\url{https://doi.org/doi:10.1007/978-3-540-24855-2_76}
\showDOI{\tempurl}


\bibitem[Vanneschi et~al\mbox{.}(2012)]%
        {Vanneschi2011}
\bibfield{author}{\bibinfo{person}{Leonardo Vanneschi}, \bibinfo{person}{Yuri
  Pirola}, \bibinfo{person}{Giancarlo Mauri}, \bibinfo{person}{Marco
  Tomassini}, \bibinfo{person}{Philippe Collard}, {and}
  \bibinfo{person}{Sebastien Verel}.} \bibinfo{year}{2012}\natexlab{}.
\newblock \showarticletitle{A study of the neutrality of {Boolean} function
  landscapes in genetic programming}.
\newblock \bibinfo{journal}{\emph{Theoretical Computer Science}}
  \bibinfo{volume}{425} (\bibinfo{date}{30 March} \bibinfo{year}{2012}),
  \bibinfo{pages}{34--57}.
\newblock
\showISSN{0304-3975}
\urldef\tempurl%
\url{https://doi.org/doi:10.1016/j.tcs.2011.03.011}
\showDOI{\tempurl}


\bibitem[Verstyuk and Douglas(2022)]%
        {verstyuk2022machine}
\bibfield{author}{\bibinfo{person}{Sergiy Verstyuk} {and}
  \bibinfo{person}{Michael~R Douglas}.} \bibinfo{year}{2022}\natexlab{}.
\newblock \showarticletitle{Machine learning the gravity equation for
  international trade}.
\newblock \bibinfo{journal}{\emph{Available at SSRN 4053795}}
  (\bibinfo{year}{2022}).
\newblock


\bibitem[Virgolin and Pissis(2022)]%
        {DBLP:journals/tmlr/VirgolinP22}
\bibfield{author}{\bibinfo{person}{Marco Virgolin} {and}
  \bibinfo{person}{Solon~P. Pissis}.} \bibinfo{year}{2022}\natexlab{}.
\newblock \showarticletitle{Symbolic Regression is NP-hard}.
\newblock \bibinfo{journal}{\emph{Transactions on Machine Learning Research}}
  \bibinfo{volume}{2022} (\bibinfo{year}{2022}).
\newblock
\urldef\tempurl%
\url{https://openreview.net/forum?id=LTiaPxqe2e}
\showURL{%
\tempurl}


\bibitem[Virgolin et~al\mbox{.}(2020)]%
        {virgolin2020machine}
\bibfield{author}{\bibinfo{person}{Marco Virgolin}, \bibinfo{person}{Ziyuan
  Wang}, \bibinfo{person}{Tanja Alderliesten}, {and} \bibinfo{person}{Peter~AN
  Bosman}.} \bibinfo{year}{2020}\natexlab{}.
\newblock \showarticletitle{Machine learning for the prediction of
  pseudorealistic pediatric abdominal phantoms for radiation dose
  reconstruction}.
\newblock \bibinfo{journal}{\emph{Journal of Medical Imaging}}
  \bibinfo{volume}{7}, \bibinfo{number}{4} (\bibinfo{year}{2020}),
  \bibinfo{pages}{046501--046501}.
\newblock


\bibitem[Yang and Affenzeller(2023)]%
        {yang2023emo}
\bibfield{author}{\bibinfo{person}{Kaifeng Yang} {and} \bibinfo{person}{Michael
  Affenzeller}.} \bibinfo{year}{2023}\natexlab{}.
\newblock \showarticletitle{Surrogate-assisted Multi-objective Optimization via
  Genetic Programming Based Symbolic Regression}. In
  \bibinfo{booktitle}{\emph{Evolutionary Multi-Criterion Optimization}},
  \bibfield{editor}{\bibinfo{person}{Michael Emmerich},
  \bibinfo{person}{Andr{\'e} Deutz}, \bibinfo{person}{Hao Wang},
  \bibinfo{person}{Anna~V. Kononova}, \bibinfo{person}{Boris Naujoks},
  \bibinfo{person}{Ke~Li}, \bibinfo{person}{Kaisa Miettinen}, {and}
  \bibinfo{person}{Iryna Yevseyeva}} (Eds.). \bibinfo{publisher}{Springer
  Nature Switzerland}, \bibinfo{address}{Cham}, \bibinfo{pages}{176--190}.
\newblock
\showISBNx{978-3-031-27250-9}


\end{thebibliography}

\clearpage
\appendix
\section{Appendix}

\subsection{Proof}
\begin{delayedproof}{th:hilbertspace}
    To prove the theorem, it is sufficient to show the existence of a bijective mapping $\pi : \H \to \R^N$ such that $\forall f \in \mathcal{F} : \pi\br{\cl{f}} \coloneqq \br{f(\BF{x}_i)}_{i=1}^N$.
    By definition \eqref{def:eq} we see that if $g_1, g_2 \in \cl{f}$ for any $f \in \mathcal{F}$, then $\br{g_1(\BF{x}_i)}_{i=1}^N = \br{g_2(\BF{x}_i)}_{i=1}^N$.
    It means that for every $v \in \H$, there exists a single $\BF{a} \in \R^N$ such that $\pi(v) = \BF{a}$.
    Now, to demonstrate that this $\pi$ is a bijection, it is sufficient to show that for all $\BF{a} \in \R^N$ there exists a unique $v \in \H$ such that $\pi(v) = \BF{a}$.
    For any such $\BF{a} \coloneqq \br{a_i}_{i = 1}^N$, let us consider \[v \coloneqq \cl{ \forall \BF{y} \in \domain: \BF{y} \mapsto 
    \begin{cases}
         a_i, & \text{if \;} \BF{x}_i = \BF{y} \\
         0, & \text{otherwise} 
    \end{cases}}\]
    It is obvious that $\pi(v) = \BF{a}$.
    Moreover, if there exists $f \in \mathcal{F}: \pi(\cl{f}) = \BF{a}$ then $\forall i \in \Z_N: f(\BF{x}_i) = a_i$, so $f \in v$, which means that $\cl{f} = v$\footnote{In this paper, we use notation $\Z_m$ to represent an integer set $\set{1, \cdots, m}$.}. 
    It means that such $v$ is unique for every $\BF{a}$.    
\end{delayedproof}

\begin{delayedproof}{theorem:4opt}
    Mathematical induction is used here to prove Theorem~\ref{theorem:4opt}.
    We start by proving by induction that Algorithm \ref{alg:ftg} maintains the following invariant. 
    For $1 \le k \le N$, element $\cl{\Fest_{k}}$ is the closest to the element $\cl{\F}$ among all the elements in the space $\Span{\cl{v_1}, \dots, \cl{v_{k}}}$. We prove it by induction over $k$.

    For $k = 1$ the element $\cl{\Fest_1} = \inner{\cl{v_1}}{\cl{v_1}}^{-1}\inner{\cl{\F}}{\cl{v_1}} \cdot \cl{v_1}$ is the solution of LLSQ, given by Eq. \eqref{eq:prj_to_subspace}. Hence $\cl{\Fest_1}$ is closest to $\cl{\F}$ in the space $\Span{\cl{v_1}}$. The statement is proven for $k = 1$. %

    Assume that the statement is correct for all values of $k$ up to value $t$ such that $1 < k \le t < N$.
    Let us prove that it is correct for $k = t + 1$.
    Based on our assumption, $\cl{\Fest_{t}}$ is the closest to element $\cl{\F}$ among all elements in the space $S_t \coloneqq \Span{\cl{v_1}, \dots, \cl{v_{t}}}$.
    Using the projection theorem, we can conclude that $\cl{\F} - \cl{\Fest_{t}}$ is perpendicular to the space $S_t$.
    It means that if the condition in the line \ref{alg:ftg:condition} is not satisfied, then the algorithm leaves the while loop and element $\cl{v_{t+1}}$ does not lie in the space $\Span{\cl{v_1}, \dots, \cl{v_{t}}}$.
    Now it is clear that for every $p > 1$ element $\cl{v_{p}}$ does not lie in the space $\Span{\cl{v_1}, \dots, \cl{v_{p-1}}}$.
    It implies that elements $\cl{v_1}$,$ \cl{v_2}$,$ \dots$,$ \cl{v_{t+1}}$ are linearly independent. 
    Hence, the inverse of matrix $\BF{G}$, computed in line \ref{alg:ftg:gram} for $k = t + 1$, exists, and so vector $\BF{\alpha}$, computed for $k = t + 1$, exists too.
    Then $\cl{\Fest_{t+1}} \in \Span{\cl{v_1}, \cl{v_2}, \dots, \cl{v_{t+1}}}$ because $\cl{\Fest_{t+1}}$ 
    is constructed as a linear combination of the corresponding elements in the line \ref{alg:ftg:update} of Algorithm \ref{alg:ftg}.
    This vector $\BF{\alpha}$ gives the closest element to the element $\cl{\F}$ because it is constructed in lines \ref{alg:ftg:gram}, \ref{alg:ftg:alpha} according to the Eq. \eqref{eq:prj_to_subspace}.
    The transition of the induction is proven and so the statement is also proven.

    Consider iteration $t$ of Algorithm \ref{alg:ftg}. 
    As we already shown \[\Span{\cl{v_1}, \cl{v_2}, \dots, \cl{v_{t-1}}} \subset \Span{\cl{v_1}, \cl{v_2}, \dots, \cl{v_t}}.\]
    Hence, $d\br{\cl{\F}, \cl{\Fest_{t}}} \le d\br{\cl{\F}, \cl{\Fest_{t-1}}}$.
    If the equality is attained, then $\cl{\Fest_{t}} = \cl{\Fest_{t-1}}$ because both $\cl{\Fest_t}$, $\cl{\Fest_{t-1}}$ belong to the same subspace $\Span{\cl{v_1}, \cl{v_2}, \dots, \cl{v_t}}$ and the element $\cl{\Fest_t}$ closest to $\cl{F}$ in this subspace is unique.
    Hence, $\inner{\F - \Fest_{t-1}}{\cl{v_t}} = 0$, which means that such $\cl{v_t}$ would not violate the condition in line \ref{alg:ftg:condition} of Algorithm \ref{alg:ftg}, and so would not be added.
    Contradiction with the assumption that equality is attained.
    Then $d\br{\cl{\F}, \cl{\Fest_{t}}} < d\br{\cl{\F}, \cl{\Fest_{t-1}}}$, which implies that $\mathcal{L}\br{\Fest_t} < \mathcal{L}\br{\Fest_{t-1}}.$
    It proves the first statement of the theorem.

    If $\mathcal{H}$ is $N$-dimensional, then $\Span{\cl{v_1}, \cl{v_2}, \dots, \cl{v_N}} = \mathcal{H}.$
    Hence, for function $\Fest_N:$ %
    $d\br{\cl{\F}, \cl{\Fest_N}} = 0 \implies \mathcal{L}\br{\Fest_N} = 0.$
    
    The second statement is also proven. 
\end{delayedproof}

\subsection{Tables}
\begin{table}[H]
\caption{Notation}
\label{tab:notations}
\scalebox{0.8}{
\begin{adjustbox}{center}
    \begin{tabular}{c|cc}
    \toprule
    Symbol          & Definition                                & Domain   \\ \hline
    $f,g$           & A function                                & $\mathbb{X} \rightarrow \mathbb{R}$ \\
    $f_{itness}$    & A fitness function for SR problems        & $\mathbb{R}$ \\
    $\BF{x}$        & Features in training dataset              & $\mathbb{R}^n$ \\
    $\X$            & Training dataset                          & $\mathbb{R}^{nN}$ \\
    $\hat{F}$       & A SR function                            & $\mathbb{X} \rightarrow \mathbb{R}$  \\
    $U$             & Uniform distribution over the given set  & n.a. \\
    $\pi$           & Bijective mapping                       & n.a. \\
    $\mathfrak{U}$  & Unary operator space                      & $\set{u \mid u:  \mathbb{R} \rightarrow \mathbb{R} }$ \\
    $\mathfrak{B}$  & Binary operator space                     & $\set{b \mid b:  \mathbb{R}^2 \rightarrow \mathbb{R} }$ \\
    $\Pi$           & Orthogonal projector operator space       & $\set{(p_i)_{i=1}^{n} \mid p_i:  \domain \rightarrow \mathbb{R} }$ \\
    $\mathfrak{C}$  & Space of constants                        & $\subseteq \R$ \\
    $\mathbb{X}$    & Domain of variables                       & $\subset \mathbb{R}^n$ \\
    $\mathcal{F}$   & Function Space                            & $\set{f \mid f : \domain \to \R}$ \\
    $\mathbb{F}$    & Quotient space                             & $\mathcal{F} / \sim$ \\
    $\Span{x}$         & Linear span of $x$                               & n.a. \\
    \bottomrule
    \end{tabular}
\end{adjustbox}
}
\end{table}

\begin{table}[H]
  \caption{Conventional symbolic regression benchmark 
  }
  \setlength{\tabcolsep}{5pt}
  \footnotesize
  \centering
  \scalebox{0.85}{
  \begin{tabular}{ l l r l }
    \toprule
   \textbf{Problem} & \textbf{Objective Function} & \textbf{Vars} &\textbf{Training Set} \\ 
    \midrule
    koza1 & $x^4 + x^3 + x^2 + x $ & $x$ & $x_1, x_2,\dotsc,x_{20} \overset{\mathrm{i.i.d.}}{\sim} U(-1, 1)$ \\ 
    koza2 & $x^5 - 2x^3 + x $ & $x$ & $x_1, x_2,\dotsc,x_{20} \overset{\mathrm{i.i.d.}}{\sim} U(-1, 1)$ \\ 
    koza3  & $x^6 - 2x^4 + x^2$ & $x$ & $x_1, x_2,\dotsc,x_{20} \overset{\mathrm{i.i.d.}}{\sim} U(-1, 1)$ \\ 
    nguyen3 & $x^5 + x^4 + x^3 + x^2 + x $ & $x$ & $x_1, x_2,\dotsc,x_{20} \overset{\mathrm{i.i.d.}}{\sim} U(-1, 1)$ \\ 
    nguyen4 & $x^6 + x^5 + x^4 + x^3 + x^2 + x$ & $x$ & $x_1, x_2,\dotsc,x_{20} \overset{\mathrm{i.i.d.}}{\sim} U(-1, 1)$ \\ 
    nguyen5 & $\sin(x^2) \cos(x) - 1$ & $x$ & $x_1, x_2,\dotsc,x_{20} \overset{\mathrm{i.i.d.}}{\sim} U(-1, 1)$ \\ 
    nguyen6 & $\sin(x) + \sin(x + x^2 )$ & $x$ & $x_1, x_2,\dotsc,x_{20} \overset{\mathrm{i.i.d.}}{\sim} U(-1, 1)$ \\ 
    nguyen7 & $\ln(x + 1) + \ln(x^2 + 1)$ & $x$ & $x_1, x_2,\dotsc,x_{20} \overset{\mathrm{i.i.d.}}{\sim} U(0, 2)$ \\ 
    nguyen8 & $\sqrt{x}$ & $x$ & $x_1, x_2,\dotsc,x_{20} \overset{\mathrm{i.i.d.}}{\sim} U(0, 4)$ \\ 
    \bottomrule
  \end{tabular}
  }
  \label{problems_regression}
  \end{table}

\begin{table}[H]
\caption{Configuration of the GP algorithms}
\scalebox{0.7}{
\begin{adjustbox}{center}
\begin{tabular}{cc|ccc}
\toprule
\multicolumn{2}{c|}{Parameters} & (1+1)-GP & (1+$\lambda$)-GP & Canonical GP \\
\midrule
$P$                & population size          & 1         &  1                &     500         \\
$\lambda$          & offspring size          &   1       &   500               &    500          \\
$M_t$              & mutation type          & uniform subtree         & uniform subtree                 &  probabilistic subtree            \\
$M_p$              & mutation rate          &  1        &   1               &    0.1          \\
$C_t$              & crossover type          &  n.a.        &     n.a.             &   subtree crossover           \\
$C_p$              & crossover rate          & n.a.         &   n.a.               &   0.9           \\
$T$                & tournament size          &  n.a.        &  n.a.               &   2          \\
\bottomrule
\end{tabular}
\label{meta-eval-parameters}
\end{adjustbox}
}
\end{table}

\begin{table*}[]
\centering
\tiny
\caption{Results of the algorithm comparison for the problems evaluated by the number of fitness evaluations (FE) to termination.}

\scalebox{1}{
\begin{tabular}{ c l l r r r r r r r }
\toprule
\scriptsize  \textbf{Tolerance} & \textbf{Problem} & \scriptsize \textbf{Algorithm} & \scriptsize \textbf{Mean FE} & \scriptsize \textbf{SD}
& \scriptsize \textbf{SEM} & \scriptsize \textbf{1Q} & \scriptsize\textbf{Median} & \scriptsize \textbf{3Q} &  \textbf{Success}
\\ & & & & & & & & &   \textbf{rate (\%)} \\ 
\midrule
\multirow{36}{*}{\rotatebox{90}{$\sum\limits_{\BF{x} \in \X} \br{F(\BF{x}) - \Fest(\BF{x})}^2 < 10^{-2}$}}
& \multirow{ 4}{*}{koza1}
& $(1+1)$-GP & $20535.662$ & $22551.856$ & $2570.023$ & $4941.000$ & $13087.000$  & $24691.000$ & $77$  \\
& & ($1+\lambda$)-GP & $21470.880$ & $18044.348$ & $1980.624$ & $8251.000$ & $15501.000$  & $29001.000$ & $83$  \\
& & Canonical-GP & $25704.898$ & $15474.615$ & $1563.172$ & $16062.500$ & $21167.000$  & $31749.500$ & $98$  \\
& & \textbf{FTG} & $\mathbf{210.790}$ & $\mathbf{499.682}$ & $\mathbf{49.968}$ & $\mathbf{117.750}$ & $\mathbf{143.500}$  & $\mathbf{198.250}$ & $\mathbf{100}$  \\
\cline{2-10}
& \multirow{ 4}{*}{koza2}
& $(1+1)$-GP & $4988.459$ & $7525.069$ & $760.147$ & $802.250$ & $1975.000$  & $4934.500$ & $98$  \\
& & ($1+\lambda$)-GP & $7883.653$ & $8582.969$ & $867.011$ & $2501.000$ & $4751.000$  & $8376.000$ & $98$  \\
& & Canonical-GP & $40612.589$ & $24258.788$ & $2488.899$ & $16934.000$ & $38846.000$  & $60260.000$ & $95$  \\
& & \textbf{FTG} & $\mathbf{135.350}$ & $\mathbf{70.287}$ & $\mathbf{7.029}$ & $\mathbf{96.250}$ & $\mathbf{121.000}$  & $\mathbf{171.000}$ & $\mathbf{100}$  \\
\cline{2-10}
& \multirow{ 4}{*}{koza3}
& $(1+1)$-GP & $12691.833$ & $18547.045$ & $2023.649$ & $2149.250$ & $6052.500$  & $12831.250$ & $84$  \\
& & ($1+\lambda$)-GP & $18552.282$ & $18159.124$ & $2056.116$ & $6001.000$ & $12251.000$  & $21251.000$ & $78$  \\
& & Canonical-GP & $17425.927$ & $19790.600$ & $2185.506$ & $4110.500$ & $9464.000$  & $21416.000$ & $83$  \\
& & \textbf{FTG} & $\mathbf{90.090}$ & $\mathbf{46.833}$ & $\mathbf{4.683}$ & $\mathbf{59.000}$ & $\mathbf{94.000}$  & $\mathbf{118.000}$ & $\mathbf{100}$  \\
\cline{2-10}
& \multirow{ 4}{*}{nguyen3}
& $(1+1)$-GP & $24293.640$ & $25123.029$ & $2709.086$ & $7748.250$ & $14440.000$  & $29669.250$ & $86$  \\
& & ($1+\lambda$)-GP & $36565.706$ & $24355.189$ & $2641.691$ & $16001.000$ & $32501.000$  & $54001.000$ & $85$  \\
& & Canonical-GP & $32333.692$ & $14479.265$ & $1517.839$ & $21914.000$ & $28388.000$  & $40340.000$ & $92$  \\
& & \textbf{FTG} & $\mathbf{187.400}$ & $\mathbf{85.965}$ & $\mathbf{8.596}$ & $\mathbf{122.750}$ & $\mathbf{168.000}$  & $\mathbf{231.250}$ & $\mathbf{100}$  \\
\cline{2-10}
& \multirow{ 4}{*}{nguyen4}
& $(1+1)$-GP & $27746.200$ & $22266.350$ & $2571.097$ & $9702.500$ & $23248.000$  & $36785.000$ & $75$  \\
& & ($1+\lambda$)-GP & $37046.455$ & $24427.627$ & $2783.786$ & $14501.000$ & $33001.000$  & $54001.000$ & $77$  \\
& & Canonical-GP & $47868.974$ & $19633.218$ & $2252.085$ & $32870.000$ & $45818.000$  & $60509.000$ & $79$  \\
& & \textbf{FTG} & $\mathbf{251.070}$ & $\mathbf{485.937}$ & $\mathbf{48.594}$ & $\mathbf{142.000}$ & $\mathbf{190.000}$  & $\mathbf{253.000}$ & $\mathbf{100}$  \\
\cline{2-10}
& \multirow{ 4}{*}{nguyen5}
& $(1+1)$-GP & $8523.299$ & $9957.890$ & $1011.070$ & $1817.000$ & $6193.000$  & $10616.000$ & $97$  \\
& & ($1+\lambda$)-GP & $13181.412$ & $12216.936$ & $1240.442$ & $5001.000$ & $9001.000$  & $16501.000$ & $97$  \\
& & Canonical-GP & $29768.818$ & $17201.884$ & $2117.406$ & $17556.500$ & $25898.000$  & $39717.500$ & $66$  \\
& & \textbf{FTG} & $\mathbf{87.590}$ & $\mathbf{44.434}$ & $\mathbf{4.443}$ & $\mathbf{56.250}$ & $\mathbf{79.000}$  & $\mathbf{118.000}$ & $\mathbf{100}$  \\
\cline{2-10}
& \multirow{ 4}{*}{nguyen6}
& $(1+1)$-GP & $15678.733$ & $19769.373$ & $2131.786$ & $3205.250$ & $7522.000$  & $20483.500$ & $86$  \\
& & ($1+\lambda$)-GP & $22548.619$ & $19765.082$ & $2156.547$ & $7251.000$ & $17501.000$  & $29126.000$ & $84$  \\
& & Canonical-GP & $19346.990$ & $13967.329$ & $1418.167$ & $10460.000$ & $14942.000$  & $23906.000$ & $97$  \\
& & \textbf{FTG} & $\mathbf{153.590}$ & $\mathbf{221.710}$ & $\mathbf{22.171}$ & $\mathbf{97.000}$ & $\mathbf{122.500}$  & $\mathbf{168.000}$ & $\mathbf{100}$  \\
\cline{2-10}
& \multirow{ 4}{*}{nguyen7}
& $(1+1)$-GP & $12280.054$ & $18832.320$ & $1952.821$ & $2204.000$ & $4458.000$  & $13355.000$ & $93$  \\
& & ($1+\lambda$)-GP & $17016.306$ & $15267.755$ & $1542.276$ & $6626.000$ & $13001.000$  & $22001.000$ & $98$  \\
& & Canonical-GP & $27446.780$ & $12216.519$ & $1221.652$ & $18428.000$ & $24653.000$  & $33866.000$ & $100$  \\
& & \textbf{FTG} & $\mathbf{48.760}$ & $\mathbf{30.594}$ & $\mathbf{3.059}$ & $\mathbf{30.000}$ & $\mathbf{44.000}$  & $\mathbf{61.000}$ & $\mathbf{100}$  \\
\cline{2-10}
& \multirow{ 4}{*}{nguyen8}
& $(1+1)$-GP & $17749.125$ & $21108.855$ & $2250.211$ & $4811.000$ & $10081.500$  & $21978.750$ & $88$  \\
& & ($1+\lambda$)-GP & $19271.833$ & $17180.099$ & $1753.437$ & $7876.000$ & $12751.000$  & $27626.000$ & $96$  \\
& & Canonical-GP & $52970.128$ & $25733.585$ & $3753.629$ & $30131.000$ & $56774.000$  & $71216.000$ & $49$  \\
& & \textbf{FTG} & $\mathbf{68.520}$ & $\mathbf{35.965}$ & $\mathbf{3.597}$ & $\mathbf{44.000}$ & $\mathbf{60.000}$  & $\mathbf{94.250}$ & $\mathbf{100}$  \\

\hline

\multirow{36}{*}{\rotatebox{90}{$\sum\limits_{\BF{x} \in \X} \br{F(\BF{x}) - \Fest(\BF{x})}^2 < 10^{-1}$}}
& \multirow{ 4}{*}{koza1}
& $(1+1)$-GP & $10031.772$ & $19916.297$ & $2076.417$ & $991.250$ & $2412.500$  & $6353.000$ & $92$  \\
& & ($1+\lambda$)-GP & $9141.625$ & $11727.750$ & $1196.958$ & $3501.000$ & $5501.000$  & $9126.000$ & $96$  \\
& & Canonical-GP & $17713.697$ & $8298.540$ & $834.035$ & $12452.000$ & $16436.000$  & $20171.000$ & $100$  \\
& & \textbf{FTG} & $\mathbf{105.920}$ & $\mathbf{49.117}$ & $\mathbf{4.912}$ & $\mathbf{75.750}$ & $\mathbf{99.000}$  & $\mathbf{127.750}$ & $\mathbf{100}$  \\
\cline{2-10}
& \multirow{ 4}{*}{koza2}
& $(1+1)$-GP & $622.031$ & $737.291$ & $74.478$ & $216.250$ & $383.000$  & $853.500$ & $98$  \\
& & ($1+\lambda$)-GP & $2216.000$ & $2657.306$ & $265.731$ & $1001.000$ & $1501.000$  & $2501.000$ & $100$  \\
& & Canonical-GP & $2601.560$ & $1674.910$ & $167.491$ & $1496.000$ & $2492.000$  & $3488.000$ & $100$  \\
& & \textbf{FTG} & $\mathbf{78.540}$ & $\mathbf{44.565}$ & $\mathbf{4.456}$ & $\mathbf{45.000}$ & $\mathbf{77.000}$  & $\mathbf{98.000}$ & $\mathbf{100}$  \\
\cline{2-10}
& \multirow{ 4}{*}{koza3}
& $(1+1)$-GP & $77.130$ & $102.375$ & $10.237$ & $21.000$ & $37.000$  & $67.000$ & $100$  \\
& & ($1+\lambda$)-GP & $621.000$ & $224.944$ & $22.494$ & $501.000$ & $501.000$  & $501.000$ & $100$  \\
& & Canonical-GP & $515.245$ & $85.788$ & $8.666$ & $500.000$ & $500.000$  & $500.000$ & $100$  \\
& & \textbf{FTG} & $\mathbf{3.000}$ & $\mathbf{0.000}$ & $\mathbf{0.000}$ & $\mathbf{3.000}$ & $\mathbf{3.000}$  & $\mathbf{3.000}$ & $\mathbf{100}$  \\
\cline{2-10}
& \multirow{ 4}{*}{nguyen3}
& $(1+1)$-GP & $5233.250$ & $9750.193$ & $995.125$ & $1696.250$ & $3073.500$  & $6071.500$ & $96$  \\
& & ($1+\lambda$)-GP & $13332.633$ & $13609.710$ & $1374.788$ & $5501.000$ & $9251.000$  & $14876.000$ & $98$  \\
& & Canonical-GP & $20242.143$ & $6818.682$ & $688.791$ & $15066.500$ & $19424.000$  & $23906.000$ & $100$  \\
& & \textbf{FTG} & $\mathbf{128.330}$ & $\mathbf{67.033}$ & $\mathbf{6.703}$ & $\mathbf{96.750}$ & $\mathbf{121.000}$  & $\mathbf{166.500}$ & $\mathbf{100}$  \\
\cline{2-10}
& \multirow{ 4}{*}{nguyen4}
& $(1+1)$-GP & $9457.990$ & $12994.070$ & $1326.202$ & $2262.000$ & $4374.500$  & $10826.750$ & $96$  \\
& & ($1+\lambda$)-GP & $13444.299$ & $12402.169$ & $1259.249$ & $6001.000$ & $9501.000$  & $17501.000$ & $97$  \\
& & Canonical-GP & $25303.613$ & $9637.023$ & $999.313$ & $19424.000$ & $24902.000$  & $29882.000$ & $96$  \\
& & \textbf{FTG} & $\mathbf{143.100}$ & $\mathbf{74.536}$ & $\mathbf{7.454}$ & $\mathbf{98.500}$ & $\mathbf{122.500}$  & $\mathbf{168.500}$ & $\mathbf{100}$  \\
\cline{2-10}
& \multirow{ 4}{*}{nguyen5}
& $(1+1)$-GP & $1278.020$ & $1815.722$ & $182.487$ & $191.000$ & $468.000$  & $1682.500$ & $99$  \\
& & ($1+\lambda$)-GP & $2771.000$ & $2831.448$ & $283.145$ & $1501.000$ & $2001.000$  & $3501.000$ & $100$  \\
& & Canonical-GP & $2666.300$ & $3414.388$ & $341.439$ & $1371.500$ & $1994.000$  & $2990.000$ & $100$  \\
& & \textbf{FTG} & $\mathbf{37.080}$ & $\mathbf{24.411}$ & $\mathbf{2.441}$ & $\mathbf{19.000}$ & $\mathbf{30.500}$  & $\mathbf{58.000}$ & $\mathbf{100}$  \\
\cline{2-10}
& \multirow{ 4}{*}{nguyen6}
& $(1+1)$-GP & $3683.396$ & $6924.684$ & $706.748$ & $657.250$ & $1480.000$  & $3616.250$ & $96$  \\
& & ($1+\lambda$)-GP & $6517.854$ & $8730.500$ & $925.431$ & $2501.000$ & $3501.000$  & $7001.000$ & $89$  \\
& & Canonical-GP & $10908.200$ & $8970.776$ & $897.078$ & $6476.000$ & $8966.000$  & $13074.500$ & $100$  \\
& & \textbf{FTG} & $\mathbf{73.630}$ & $\mathbf{37.696}$ & $\mathbf{3.770}$ & $\mathbf{44.000}$ & $\mathbf{60.500}$  & $\mathbf{97.000}$ & $\mathbf{100}$  \\
\cline{2-10}
& \multirow{ 4}{*}{nguyen7}
& $(1+1)$-GP & $563.640$ & $579.752$ & $57.975$ & $140.250$ & $366.500$  & $774.500$ & $100$  \\
& & ($1+\lambda$)-GP & $1391.000$ & $829.397$ & $82.940$ & $1001.000$ & $1001.000$  & $1501.000$ & $100$  \\
& & Canonical-GP & $2093.600$ & $1372.891$ & $137.289$ & $998.000$ & $1994.000$  & $2990.000$ & $100$  \\
& & \textbf{FTG} & $\mathbf{35.290}$ & $\mathbf{24.698}$ & $\mathbf{2.470}$ & $\mathbf{17.250}$ & $\mathbf{30.500}$  & $\mathbf{46.250}$ & $\mathbf{100}$  \\
\cline{2-10}
& \multirow{ 4}{*}{nguyen8}
& $(1+1)$-GP & $4221.566$ & $7441.241$ & $747.873$ & $820.000$ & $1789.000$  & $3849.000$ & $99$  \\
& & ($1+\lambda$)-GP & $5556.000$ & $5739.728$ & $573.973$ & $2001.000$ & $4001.000$  & $6501.000$ & $100$  \\
& & Canonical-GP & $21288.733$ & $24501.236$ & $2582.657$ & $3986.000$ & $8966.000$  & $34986.500$ & $94$  \\
& & \textbf{FTG} & $\mathbf{40.080}$ & $\mathbf{25.271}$ & $\mathbf{2.527}$ & $\mathbf{20.000}$ & $\mathbf{32.000}$  & $\mathbf{59.000}$ & $\mathbf{100}$  \\

\hline

\multirow{36}{*}{\rotatebox{90}{$\sum\limits_{\BF{x} \in \X} \br{F(\BF{x}) - \Fest(\BF{x})}^2 < 1$}}
& \multirow{ 4}{*}{koza1}
& $(1+1)$-GP & $1535.396$ & $7268.502$ & $741.838$ & $162.750$ & $435.500$  & $818.000$ & $96$  \\
& & ($1+\lambda$)-GP & $2192.919$ & $1419.039$ & $142.619$ & $1501.000$ & $2001.000$  & $2501.000$ & $99$  \\
& & Canonical-GP & $4544.364$ & $2427.620$ & $243.985$ & $2741.000$ & $4484.000$  & $5978.000$ & $100$  \\
& & \textbf{FTG} & $\mathbf{57.780}$ & $\mathbf{36.299}$ & $\mathbf{3.630}$ & $\mathbf{31.000}$ & $\mathbf{45.000}$  & $\mathbf{77.000}$ & $\mathbf{100}$  \\
\cline{2-10}
& \multirow{ 4}{*}{koza2}
& $(1+1)$-GP & $16.420$ & $26.715$ & $2.672$ & $5.000$ & $11.000$  & $18.500$ & $100$  \\
& & ($1+\lambda$)-GP & $456.000$ & $143.091$ & $14.309$ & $501.000$ & $501.000$  & $501.000$ & $100$  \\
& & Canonical-GP & $500.000$ & $0.000$ & $0.000$ & $500.000$ & $500.000$  & $500.000$ & $100$  \\
& & \textbf{FTG} & $\mathbf{3.250}$ & $\mathbf{1.915}$ & $\mathbf{0.192}$ & $\mathbf{3.000}$ & $\mathbf{3.000}$  & $\mathbf{3.000}$ & $\mathbf{100}$  \\
\cline{2-10}
& \multirow{ 4}{*}{koza3}
& $(1+1)$-GP & $13.520$ & $14.876$ & $1.488$ & $4.000$ & $8.500$  & $16.250$ & $100$  \\
& & ($1+\lambda$)-GP & $446.000$ & $156.445$ & $15.644$ & $501.000$ & $501.000$  & $501.000$ & $100$  \\
& & Canonical-GP & $500.000$ & $0.000$ & $0.000$ & $500.000$ & $500.000$  & $500.000$ & $100$  \\
& & \textbf{FTG} & $\mathbf{3.000}$ & $\mathbf{0.000}$ & $\mathbf{0.000}$ & $\mathbf{3.000}$ & $\mathbf{3.000}$  & $\mathbf{3.000}$ & $\mathbf{100}$  \\
\cline{2-10}
& \multirow{ 4}{*}{nguyen3}
& $(1+1)$-GP & $862.610$ & $1099.659$ & $109.966$ & $262.000$ & $539.000$  & $1077.750$ & $100$  \\
& & ($1+\lambda$)-GP & $3051.000$ & $1854.050$ & $185.405$ & $1501.000$ & $2501.000$  & $4001.000$ & $100$  \\
& & Canonical-GP & $8442.592$ & $4089.553$ & $413.107$ & $5604.500$ & $8468.000$  & $10958.000$ & $100$  \\
& & \textbf{FTG} & $\mathbf{72.450}$ & $\mathbf{42.992}$ & $\mathbf{4.299}$ & $\mathbf{43.750}$ & $\mathbf{59.500}$  & $\mathbf{96.250}$ & $\mathbf{100}$  \\
\cline{2-10}
& \multirow{ 4}{*}{nguyen4}
& $(1+1)$-GP & $1974.770$ & $4756.862$ & $475.686$ & $409.750$ & $677.500$  & $1639.250$ & $100$  \\
& & ($1+\lambda$)-GP & $3456.000$ & $3029.930$ & $302.993$ & $2001.000$ & $3001.000$  & $4001.000$ & $100$  \\
& & Canonical-GP & $9780.438$ & $5450.162$ & $556.255$ & $5480.000$ & $9713.000$  & $13448.000$ & $100$  \\
& & \textbf{FTG} & $\mathbf{81.910}$ & $\mathbf{50.198}$ & $\mathbf{5.020}$ & $\mathbf{45.000}$ & $\mathbf{70.000}$  & $\mathbf{100.500}$ & $\mathbf{100}$  \\
\cline{2-10}
& \multirow{ 4}{*}{nguyen5}
& $(1+1)$-GP & $35.910$ & $35.989$ & $3.599$ & $10.750$ & $23.500$  & $49.500$ & $100$  \\
& & ($1+\lambda$)-GP & $521.000$ & $140.000$ & $14.000$ & $501.000$ & $501.000$  & $501.000$ & $100$  \\
& & Canonical-GP & $500.000$ & $0.000$ & $0.000$ & $500.000$ & $500.000$  & $500.000$ & $100$  \\
& & \textbf{FTG} & $\mathbf{3.000}$ & $\mathbf{0.000}$ & $\mathbf{0.000}$ & $\mathbf{3.000}$ & $\mathbf{3.000}$  & $\mathbf{3.000}$ & $\mathbf{100}$  \\
\cline{2-10}
& \multirow{ 4}{*}{nguyen6}
& $(1+1)$-GP & $766.505$ & $3743.581$ & $376.244$ & $94.000$ & $180.000$  & $373.000$ & $99$  \\
& & ($1+\lambda$)-GP & $2073.165$ & $5258.273$ & $533.897$ & $1001.000$ & $1501.000$  & $1501.000$ & $97$  \\
& & Canonical-GP & $2013.920$ & $1534.810$ & $153.481$ & $873.500$ & $1496.000$  & $2990.000$ & $100$  \\
& & \textbf{FTG} & $\mathbf{35.360}$ & $\mathbf{32.344}$ & $\mathbf{3.234}$ & $\mathbf{11.000}$ & $\mathbf{21.000}$  & $\mathbf{46.000}$ & $\mathbf{100}$  \\
\cline{2-10}
& \multirow{ 4}{*}{nguyen7}
& $(1+1)$-GP & $148.160$ & $183.455$ & $18.345$ & $43.750$ & $85.500$  & $184.500$ & $100$  \\
& & ($1+\lambda$)-GP & $756.000$ & $278.343$ & $27.834$ & $501.000$ & $501.000$  & $1001.000$ & $100$  \\
& & Canonical-GP & $614.540$ & $262.149$ & $26.215$ & $500.000$ & $500.000$  & $500.000$ & $100$  \\
& & \textbf{FTG} & $\mathbf{21.790}$ & $\mathbf{14.384}$ & $\mathbf{1.438}$ & $\mathbf{10.000}$ & $\mathbf{19.000}$  & $\mathbf{31.000}$ & $\mathbf{100}$  \\
\cline{2-10}
& \multirow{ 4}{*}{nguyen8}
& $(1+1)$-GP & $447.690$ & $597.084$ & $59.708$ & $99.500$ & $234.000$  & $514.250$ & $100$  \\
& & ($1+\lambda$)-GP & $1351.000$ & $912.414$ & $91.241$ & $1001.000$ & $1001.000$  & $1501.000$ & $100$  \\
& & Canonical-GP & $1076.632$ & $874.034$ & $89.674$ & $500.000$ & $500.000$  & $1496.000$ & $100$  \\
& & \textbf{FTG} & $\mathbf{19.590}$ & $\mathbf{13.080}$ & $\mathbf{1.308}$ & $\mathbf{10.000}$ & $\mathbf{12.000}$  & $\mathbf{30.000}$ & $\mathbf{100}$  \\
\bottomrule
\end{tabular}
}
\label{tab:rst2}
\end{table*}

\begin{table*}[h!]
\centering
\tiny
\caption{Results of the algorithm comparison for the problems evaluated by the number of fitness evaluations (FE) to termination.}
\begin{tabular}{ c | c}
\scalebox{0.76}{
\begin{tabular}{ c l l r r r r r r r }
\toprule
\scriptsize  \textbf{Tolerance} & \textbf{Problem} & \scriptsize \textbf{Algorithm} & \scriptsize \textbf{Mean FE} & \scriptsize \textbf{SD}
& \scriptsize \textbf{SEM} & \scriptsize \textbf{1Q} & \scriptsize\textbf{Median} & \scriptsize \textbf{3Q} &  \textbf{Success}
\\ & & & & & & & & &   \textbf{rate (\%)} \\ 
\midrule
\multirow{36}{*}{\rotatebox{90}{$\sum\limits_{\BF{x} \in \X} \br{F(\BF{x}) - \Fest(\BF{x})}^2 < 10^{-8}$}}
& \multirow{ 4}{*}{koza1}
& $(1+1)$-GP & $\infty$ & $0$ & $0$ & $\infty$ & $\infty$  & $\infty$ & $0$  \\
& & ($1+\lambda$)-GP & $14438.500$ & $25864.355$ & $9144.430$ & $2251.000$ & $3751.000$  & $7876.000$ & $8$  \\
& & Canonical-GP & $22230.909$ & $9264.695$ & $1055.810$ & $15938.000$ & $19922.000$  & $25898.000$ & $77$  \\
& & \textbf{FTG} & $\mathbf{1288.330}$ & $\mathbf{2343.521}$ & $\mathbf{237.948}$ & $\mathbf{533.000}$ & $\mathbf{597.000}$  & $\mathbf{920.000}$ & $\mathbf{97}$  \\
\cline{2-10}
& \multirow{ 4}{*}{koza2}
& $(1+1)$-GP & $\infty$ & $0$ & $0$ & $\infty$ & $\infty$  & $\infty$ & $0$  \\
& & ($1+\lambda$)-GP & $\infty$ & $0$ & $0$ & $\infty$ & $\infty$  & $\infty$ & $0$  \\
& & Canonical-GP & $90140.000$ & $0.000$ & $0.000$ & $90140.000$ & $90140.000$  & $90140.000$ & $1$  \\
& & \textbf{FTG} & $\mathbf{1561.629}$ & $\mathbf{3352.298}$ & $\mathbf{340.374}$ & $\mathbf{534.000}$ & $\mathbf{626.000}$  & $\mathbf{1038.000}$ & $\mathbf{97}$  \\
\cline{2-10}
& \multirow{ 4}{*}{koza3}
& $(1+1)$-GP & $\infty$ & $0$ & $0$ & $\infty$ & $\infty$  & $\infty$ & $0$  \\
& & ($1+\lambda$)-GP & $\infty$ & $0$ & $0$ & $\infty$ & $\infty$  & $\infty$ & $0$  \\
& & Canonical-GP & $\infty$ & $0$ & $0$ & $\infty$ & $\infty$  & $\infty$ & $0$  \\
& & \textbf{FTG} & $\mathbf{1515.350}$ & $\mathbf{3482.339}$ & $\mathbf{348.234}$ & $\mathbf{542.500}$ & $\mathbf{633.000}$  & $\mathbf{999.250}$ & $\mathbf{100}$  \\
\cline{2-10}
& \multirow{ 4}{*}{nguyen3}
& $(1+1)$-GP & $\infty$ & $0$ & $0$ & $\infty$ & $\infty$  & $\infty$ & $0$  \\
& & ($1+\lambda$)-GP & $7001.000$ & $0.000$ & $0.000$ & $7001.000$ & $7001.000$  & $7001.000$ & $1$  \\
& & Canonical-GP & $26997.358$ & $10211.628$ & $1402.675$ & $20420.000$ & $24404.000$  & $32372.000$ & $54$  \\
& & \textbf{FTG} & $\mathbf{1962.102}$ & $\mathbf{4402.288}$ & $\mathbf{444.698}$ & $\mathbf{539.500}$ & $\mathbf{660.500}$  & $\mathbf{1235.750}$ & $\mathbf{98}$  \\
\cline{2-10}
& \multirow{ 4}{*}{nguyen4}
& $(1+1)$-GP & $\infty$ & $0$ & $0$ & $\infty$ & $\infty$  & $\infty$ & $0$  \\
& & ($1+\lambda$)-GP & $\infty$ & $0$ & $0$ & $\infty$ & $\infty$  & $\infty$ & $0$  \\
& & Canonical-GP & $37827.364$ & $13533.309$ & $2885.311$ & $29259.500$ & $35360.000$  & $45320.000$ & $22$  \\
& & \textbf{FTG} & $\mathbf{2752.073}$ & $\mathbf{8105.071}$ & $\mathbf{827.220}$ & $\mathbf{538.000}$ & $\mathbf{691.500}$  & $\mathbf{1201.000}$ & $\mathbf{96}$  \\
\cline{2-10}
& \multirow{ 4}{*}{nguyen5}
& $(1+1)$-GP & $\infty$ & $0$ & $0$ & $\infty$ & $\infty$  & $\infty$ & $0$  \\
& & ($1+\lambda$)-GP & $\infty$ & $0$ & $0$ & $\infty$ & $\infty$  & $\infty$ & $0$  \\
& & Canonical-GP & $18428.000$ & $0.000$ & $0.000$ & $18428.000$ & $18428.000$  & $18428.000$ & $1$  \\
& & \textbf{FTG} & $\mathbf{1393.930}$ & $\mathbf{3897.976}$ & $\mathbf{389.798}$ & $\mathbf{528.750}$ & $\mathbf{628.000}$  & $\mathbf{834.750}$ & $\mathbf{100}$  \\
\cline{2-10}
& \multirow{ 4}{*}{nguyen6}
& $(1+1)$-GP & $2192.000$ & $1468.974$ & $656.945$ & $936.000$ & $1443.000$  & $3364.000$ & $5$  \\
& & ($1+\lambda$)-GP & $2701.000$ & $678.233$ & $303.315$ & $2501.000$ & $2501.000$  & $2501.000$ & $5$  \\
& & Canonical-GP & $20408.933$ & $14429.970$ & $1521.052$ & $10958.000$ & $16934.000$  & $24653.000$ & $90$  \\
& & \textbf{FTG} & $\mathbf{2430.592}$ & $\mathbf{9119.704}$ & $\mathbf{921.229}$ & $\mathbf{527.750}$ & $\mathbf{605.000}$  & $\mathbf{989.000}$ & $\mathbf{98}$  \\
\cline{2-10}
& \multirow{ 4}{*}{nguyen7}
& $(1+1)$-GP & $\infty$ & $0$ & $0$ & $\infty$ & $\infty$  & $\infty$ & $0$  \\
& & ($1+\lambda$)-GP & $\infty$ & $0$ & $0$ & $\infty$ & $\infty$  & $\infty$ & $0$  \\
& & Canonical-GP & $\infty$ & $0$ & $0$ & $\infty$ & $\infty$  & $\infty$ & $0$  \\
& & \textbf{FTG} & $\mathbf{1930.240}$ & $\mathbf{6211.331}$ & $\mathbf{621.133}$ & $\mathbf{552.250}$ & $\mathbf{670.000}$  & $\mathbf{911.500}$ & $\mathbf{100}$  \\
\cline{2-10}
& \multirow{ 4}{*}{nguyen8}
& $(1+1)$-GP & $\infty$ & $0$ & $0$ & $\infty$ & $\infty$  & $\infty$ & $0$  \\
& & ($1+\lambda$)-GP & $\infty$ & $0$ & $0$ & $\infty$ & $\infty$  & $\infty$ & $0$  \\
& & Canonical-GP & $\infty$ & $0$ & $0$ & $\infty$ & $\infty$  & $\infty$ & $0$  \\
& & \textbf{FTG} & $\mathbf{1408.560}$ & $\mathbf{5689.417}$ & $\mathbf{568.942}$ & $\mathbf{478.000}$ & $\mathbf{559.500}$  & $\mathbf{718.500}$ & $\mathbf{100}$  \\

\hline

\multirow{36}{*}{\rotatebox{90}{$\sum\limits_{\BF{x} \in \X} \br{F(\BF{x}) - \Fest(\BF{x})}^2 < 10^{-7}$}}
& \multirow{ 4}{*}{koza1}
& $(1+1)$-GP & $\infty$ & $0$ & $0$ & $\infty$ & $\infty$  & $\infty$ & $0$  \\
& & ($1+\lambda$)-GP & $14556.556$ & $22807.704$ & $7602.568$ & $2501.000$ & $4501.000$  & $15001.000$ & $9$  \\
& & Canonical-GP & $22217.974$ & $9270.688$ & $1056.493$ & $15938.000$ & $19922.000$  & $25898.000$ & $77$  \\
& & \textbf{FTG} & $\mathbf{971.580}$ & $\mathbf{2028.005}$ & $\mathbf{202.800}$ & $\mathbf{494.750}$ & $\mathbf{550.500}$  & $\mathbf{693.500}$ & $\mathbf{100}$  \\
\cline{2-10}
& \multirow{ 4}{*}{koza2}
& $(1+1)$-GP & $\infty$ & $0$ & $0$ & $\infty$ & $\infty$  & $\infty$ & $0$  \\
& & ($1+\lambda$)-GP & $\infty$ & $0$ & $0$ & $\infty$ & $\infty$  & $\infty$ & $0$  \\
& & Canonical-GP & $90140.000$ & $0.000$ & $0.000$ & $90140.000$ & $90140.000$  & $90140.000$ & $1$  \\
& & \textbf{FTG} & $\mathbf{968.990}$ & $\mathbf{1257.073}$ & $\mathbf{126.984}$ & $\mathbf{508.500}$ & $\mathbf{544.000}$  & $\mathbf{677.000}$ & $\mathbf{98}$  \\
\cline{2-10}
& \multirow{ 4}{*}{koza3}
& $(1+1)$-GP & $\infty$ & $0$ & $0$ & $\infty$ & $\infty$  & $\infty$ & $0$  \\
& & ($1+\lambda$)-GP & $\infty$ & $0$ & $0$ & $\infty$ & $\infty$  & $\infty$ & $0$  \\
& & Canonical-GP & $\infty$ & $0$ & $0$ & $\infty$ & $\infty$  & $\infty$ & $0$  \\
& & \textbf{FTG} & $\mathbf{1053.150}$ & $\mathbf{2248.115}$ & $\mathbf{224.812}$ & $\mathbf{511.000}$ & $\mathbf{561.000}$  & $\mathbf{785.500}$ & $\mathbf{100}$  \\
\cline{2-10}
& \multirow{ 4}{*}{nguyen3}
& $(1+1)$-GP & $\infty$ & $0$ & $0$ & $\infty$ & $\infty$  & $\infty$ & $0$  \\
& & ($1+\lambda$)-GP & $7001.000$ & $0.000$ & $0.000$ & $7001.000$ & $7001.000$  & $7001.000$ & $1$  \\
& & Canonical-GP & $26997.358$ & $10211.628$ & $1402.675$ & $20420.000$ & $24404.000$  & $32372.000$ & $54$  \\
& & \textbf{FTG} & $\mathbf{1410.818}$ & $\mathbf{2961.360}$ & $\mathbf{297.628}$ & $\mathbf{503.500}$ & $\mathbf{567.000}$  & $\mathbf{985.500}$ & $\mathbf{99}$  \\
\cline{2-10}
& \multirow{ 4}{*}{nguyen4}
& $(1+1)$-GP & $\infty$ & $0$ & $0$ & $\infty$ & $\infty$  & $\infty$ & $0$  \\
& & ($1+\lambda$)-GP & $\infty$ & $0$ & $0$ & $\infty$ & $\infty$  & $\infty$ & $0$  \\
& & Canonical-GP & $37827.364$ & $13533.309$ & $2885.311$ & $29259.500$ & $35360.000$  & $45320.000$ & $22$  \\
& & \textbf{FTG} & $\mathbf{2078.031}$ & $\mathbf{5165.032}$ & $\mathbf{521.747}$ & $\mathbf{514.000}$ & $\mathbf{604.500}$  & $\mathbf{1032.750}$ & $\mathbf{98}$  \\
\cline{2-10}
& \multirow{ 4}{*}{nguyen5}
& $(1+1)$-GP & $\infty$ & $0$ & $0$ & $\infty$ & $\infty$  & $\infty$ & $0$  \\
& & ($1+\lambda$)-GP & $\infty$ & $0$ & $0$ & $\infty$ & $\infty$  & $\infty$ & $0$  \\
& & Canonical-GP & $15440.000$ & $0.000$ & $0.000$ & $15440.000$ & $15440.000$  & $15440.000$ & $1$  \\
& & \textbf{FTG} & $\mathbf{1158.660}$ & $\mathbf{3782.549}$ & $\mathbf{378.255}$ & $\mathbf{467.750}$ & $\mathbf{537.000}$  & $\mathbf{648.250}$ & $\mathbf{100}$  \\
\cline{2-10}
& \multirow{ 4}{*}{nguyen6}
& $(1+1)$-GP & $2192.000$ & $1468.974$ & $656.945$ & $936.000$ & $1443.000$  & $3364.000$ & $5$  \\
& & ($1+\lambda$)-GP & $2701.000$ & $678.233$ & $303.315$ & $2501.000$ & $2501.000$  & $2501.000$ & $5$  \\
& & Canonical-GP & $20392.333$ & $14420.492$ & $1520.053$ & $10958.000$ & $16934.000$  & $24653.000$ & $90$  \\
& & \textbf{FTG} & $\mathbf{1056.337}$ & $\mathbf{1799.477}$ & $\mathbf{181.775}$ & $\mathbf{493.250}$ & $\mathbf{556.000}$  & $\mathbf{756.000}$ & $\mathbf{98}$  \\
\cline{2-10}
& \multirow{ 4}{*}{nguyen7}
& $(1+1)$-GP & $\infty$ & $0$ & $0$ & $\infty$ & $\infty$  & $\infty$ & $0$  \\
& & ($1+\lambda$)-GP & $\infty$ & $0$ & $0$ & $\infty$ & $\infty$  & $\infty$ & $0$  \\
& & Canonical-GP & $\infty$ & $0$ & $0$ & $\infty$ & $\infty$  & $\infty$ & $0$  \\
& & \textbf{FTG} & $\mathbf{950.410}$ & $\mathbf{2804.880}$ & $\mathbf{280.488}$ & $\mathbf{403.250}$ & $\mathbf{504.000}$  & $\mathbf{609.750}$ & $\mathbf{100}$  \\
\cline{2-10}
& \multirow{ 4}{*}{nguyen8}
& $(1+1)$-GP & $90563.000$ & $0.000$ & $0.000$ & $90563.000$ & $90563.000$  & $90563.000$ & $1$  \\
& & ($1+\lambda$)-GP & $\infty$ & $0$ & $0$ & $\infty$ & $\infty$  & $\infty$ & $0$  \\
& & Canonical-GP & $\infty$ & $0$ & $0$ & $\infty$ & $\infty$  & $\infty$ & $0$  \\
& & \textbf{FTG} & $\mathbf{959.320}$ & $\mathbf{5207.160}$ & $\mathbf{520.716}$ & $\mathbf{323.000}$ & $\mathbf{432.500}$  & $\mathbf{509.750}$ & $\mathbf{100}$  \\

\hline

\multirow{36}{*}{\rotatebox{90}{$\sum\limits_{\BF{x} \in \X} \br{F(\BF{x}) - \Fest(\BF{x})}^2 < 10^{-6}$}}
& \multirow{ 4}{*}{koza1}
& $(1+1)$-GP & $47988.000$ & $36922.000$ & $26107.797$ & $29527.000$ & $47988.000$  & $66449.000$ & $2$  \\
& & ($1+\lambda$)-GP & $19151.000$ & $26725.503$ & $8451.346$ & $2626.000$ & $5001.000$  & $16126.000$ & $10$  \\
& & Canonical-GP & $22166.234$ & $9258.335$ & $1055.085$ & $15938.000$ & $19922.000$  & $25400.000$ & $77$  \\
& & \textbf{FTG} & $\mathbf{839.220}$ & $\mathbf{1998.643}$ & $\mathbf{199.864}$ & $\mathbf{403.750}$ & $\mathbf{499.000}$  & $\mathbf{559.250}$ & $\mathbf{100}$  \\
\cline{2-10}
& \multirow{ 4}{*}{koza2}
& $(1+1)$-GP & $84218.000$ & $13203.938$ & $7623.297$ & $77494.500$ & $88715.000$  & $93190.000$ & $3$  \\
& & ($1+\lambda$)-GP & $90001.000$ & $2500.000$ & $1767.767$ & $88751.000$ & $90001.000$  & $91251.000$ & $2$  \\
& & Canonical-GP & $90140.000$ & $0.000$ & $0.000$ & $90140.000$ & $90140.000$  & $90140.000$ & $1$  \\
& & \textbf{FTG} & $\mathbf{802.969}$ & $\mathbf{1206.168}$ & $\mathbf{121.841}$ & $\mathbf{419.750}$ & $\mathbf{506.500}$  & $\mathbf{593.750}$ & $\mathbf{98}$  \\
\cline{2-10}
& \multirow{ 4}{*}{koza3}
& $(1+1)$-GP & $70642.500$ & $22277.500$ & $15752.571$ & $59503.750$ & $70642.500$  & $81781.250$ & $2$  \\
& & ($1+\lambda$)-GP & $78501.000$ & $0.000$ & $0.000$ & $78501.000$ & $78501.000$  & $78501.000$ & $1$  \\
& & Canonical-GP & $\infty$ & $0$ & $0$ & $\infty$ & $\infty$  & $\infty$ & $0$  \\
& & \textbf{FTG} & $\mathbf{885.680}$ & $\mathbf{2248.385}$ & $\mathbf{224.838}$ & $\mathbf{440.500}$ & $\mathbf{507.500}$  & $\mathbf{559.250}$ & $\mathbf{100}$  \\
\cline{2-10}
& \multirow{ 4}{*}{nguyen3}
& $(1+1)$-GP & $\infty$ & $0$ & $0$ & $\infty$ & $\infty$  & $\infty$ & $0$  \\
& & ($1+\lambda$)-GP & $7001.000$ & $0.000$ & $0.000$ & $7001.000$ & $7001.000$  & $7001.000$ & $1$  \\
& & Canonical-GP & $26865.811$ & $10088.563$ & $1385.771$ & $20420.000$ & $24404.000$  & $32372.000$ & $54$  \\
& & \textbf{FTG} & $\mathbf{1252.354}$ & $\mathbf{2951.116}$ & $\mathbf{296.598}$ & $\mathbf{452.000}$ & $\mathbf{510.000}$  & $\mathbf{696.000}$ & $\mathbf{99}$  \\
\cline{2-10}
& \multirow{ 4}{*}{nguyen4}
& $(1+1)$-GP & $\infty$ & $0$ & $0$ & $\infty$ & $\infty$  & $\infty$ & $0$  \\
& & ($1+\lambda$)-GP & $61501.000$ & $0.000$ & $0.000$ & $61501.000$ & $61501.000$  & $61501.000$ & $1$  \\
& & Canonical-GP & $39127.478$ & $14573.055$ & $3038.692$ & $29633.000$ & $36356.000$  & $46067.000$ & $23$  \\
& & \textbf{FTG} & $\mathbf{1744.000}$ & $\mathbf{4936.152}$ & $\mathbf{498.627}$ & $\mathbf{466.250}$ & $\mathbf{522.500}$  & $\mathbf{779.000}$ & $\mathbf{98}$  \\
\cline{2-10}
& \multirow{ 4}{*}{nguyen5}
& $(1+1)$-GP & $98966.000$ & $0.000$ & $0.000$ & $98966.000$ & $98966.000$  & $98966.000$ & $1$  \\
& & ($1+\lambda$)-GP & $98001.000$ & $0.000$ & $0.000$ & $98001.000$ & $98001.000$  & $98001.000$ & $1$  \\
& & Canonical-GP & $15440.000$ & $0.000$ & $0.000$ & $15440.000$ & $15440.000$  & $15440.000$ & $1$  \\
& & \textbf{FTG} & $\mathbf{911.430}$ & $\mathbf{3709.353}$ & $\mathbf{370.935}$ & $\mathbf{338.750}$ & $\mathbf{426.000}$  & $\mathbf{525.000}$ & $\mathbf{100}$  \\
\cline{2-10}
& \multirow{ 4}{*}{nguyen6}
& $(1+1)$-GP & $20049.714$ & $32794.626$ & $12395.204$ & $1189.500$ & $3364.000$  & $19022.000$ & $7$  \\
& & ($1+\lambda$)-GP & $2701.000$ & $678.233$ & $303.315$ & $2501.000$ & $2501.000$  & $2501.000$ & $5$  \\
& & Canonical-GP & $21011.033$ & $15667.276$ & $1642.377$ & $10958.000$ & $16934.000$  & $24902.000$ & $91$  \\
& & \textbf{FTG} & $\mathbf{828.660}$ & $\mathbf{1140.343}$ & $\mathbf{114.034}$ & $\mathbf{410.000}$ & $\mathbf{502.500}$  & $\mathbf{594.250}$ & $\mathbf{100}$  \\
\cline{2-10}
& \multirow{ 4}{*}{nguyen7}
& $(1+1)$-GP & $94875.000$ & $4763.204$ & $2750.037$ & $92492.500$ & $96616.000$  & $98128.000$ & $3$  \\
& & ($1+\lambda$)-GP & $59001.000$ & $1000.000$ & $707.107$ & $58501.000$ & $59001.000$  & $59501.000$ & $2$  \\
& & Canonical-GP & $\infty$ & $0$ & $0$ & $\infty$ & $\infty$  & $\infty$ & $0$  \\
& & \textbf{FTG} & $\mathbf{420.080}$ & $\mathbf{769.980}$ & $\mathbf{76.998}$ & $\mathbf{256.250}$ & $\mathbf{336.500}$  & $\mathbf{431.750}$ & $\mathbf{100}$  \\
\cline{2-10}
& \multirow{ 4}{*}{nguyen8}
& $(1+1)$-GP & $49606.000$ & $917.000$ & $648.417$ & $49147.500$ & $49606.000$  & $50064.500$ & $2$  \\
& & ($1+\lambda$)-GP & $\infty$ & $0$ & $0$ & $\infty$ & $\infty$  & $\infty$ & $0$  \\
& & Canonical-GP & $\infty$ & $0$ & $0$ & $\infty$ & $\infty$  & $\infty$ & $0$  \\
& & \textbf{FTG} & $\mathbf{844.760}$ & $\mathbf{5214.687}$ & $\mathbf{521.469}$ & $\mathbf{210.750}$ & $\mathbf{322.500}$  & $\mathbf{431.250}$ & $\mathbf{100}$  \\

\bottomrule
\end{tabular}
} 
&
\scalebox{0.76}{
\begin{tabular}{ c l l r r r r r r r }
\toprule
\scriptsize  \textbf{Tolerance} & \textbf{Problem} & \scriptsize \textbf{Algorithm} & \scriptsize \textbf{Mean FE} & \scriptsize \textbf{SD}
& \scriptsize \textbf{SEM} & \scriptsize \textbf{1Q} & \scriptsize\textbf{Median} & \scriptsize \textbf{3Q} &  \textbf{Success}
\\ & & & & & & & & &   \textbf{rate (\%)} \\ 
\midrule
\multirow{36}{*}{\rotatebox{90}{$\sum\limits_{\BF{x} \in \X} \br{F(\BF{x}) - \Fest(\BF{x})}^2 < 10^{-5}$}}
& \multirow{ 4}{*}{koza1}
& $(1+1)$-GP & $53377.000$ & $31411.295$ & $12823.608$ & $32432.500$ & $61504.500$  & $75989.000$ & $6$  \\
& & ($1+\lambda$)-GP & $23046.455$ & $31240.602$ & $9419.396$ & $2751.000$ & $5501.000$  & $23751.000$ & $11$  \\
& & Canonical-GP & $22108.026$ & $9240.413$ & $1053.043$ & $15938.000$ & $19424.000$  & $25400.000$ & $77$  \\
& & \textbf{FTG} & $\mathbf{712.440}$ & $\mathbf{1942.562}$ & $\mathbf{194.256}$ & $\mathbf{302.750}$ & $\mathbf{399.500}$  & $\mathbf{495.750}$ & $\mathbf{100}$  \\
\cline{2-10}
& \multirow{ 4}{*}{koza2}
& $(1+1)$-GP & $64166.632$ & $23640.045$ & $5423.398$ & $47201.500$ & $69234.000$  & $84645.500$ & $19$  \\
& & ($1+\lambda$)-GP & $61143.857$ & $23647.281$ & $6320.002$ & $44376.000$ & $65251.000$  & $75126.000$ & $14$  \\
& & Canonical-GP & $84662.000$ & $1992.000$ & $1408.557$ & $83666.000$ & $84662.000$  & $85658.000$ & $2$  \\
& & \textbf{FTG} & $\mathbf{640.850}$ & $\mathbf{1050.594}$ & $\mathbf{105.059}$ & $\mathbf{345.750}$ & $\mathbf{419.000}$  & $\mathbf{502.500}$ & $\mathbf{100}$  \\
\cline{2-10}
& \multirow{ 4}{*}{koza3}
& $(1+1)$-GP & $58571.400$ & $29313.534$ & $13109.411$ & $44579.000$ & $64125.000$  & $80018.000$ & $5$  \\
& & ($1+\lambda$)-GP & $66667.667$ & $16483.999$ & $9517.041$ & $58001.000$ & $71501.000$  & $77751.000$ & $3$  \\
& & Canonical-GP & $\infty$ & $0$ & $0$ & $\infty$ & $\infty$  & $\infty$ & $0$  \\
& & \textbf{FTG} & $\mathbf{476.080}$ & $\mathbf{416.455}$ & $\mathbf{41.645}$ & $\mathbf{335.000}$ & $\mathbf{417.000}$  & $\mathbf{480.250}$ & $\mathbf{100}$  \\
\cline{2-10}
& \multirow{ 4}{*}{nguyen3}
& $(1+1)$-GP & $\infty$ & $0$ & $0$ & $\infty$ & $\infty$  & $\infty$ & $0$  \\
& & ($1+\lambda$)-GP & $49001.000$ & $24713.357$ & $12356.678$ & $45501.000$ & $61501.000$  & $65001.000$ & $4$  \\
& & Canonical-GP & $27890.000$ & $13072.610$ & $1778.957$ & $20544.500$ & $24404.000$  & $33492.500$ & $55$  \\
& & \textbf{FTG} & $\mathbf{768.830}$ & $\mathbf{1396.386}$ & $\mathbf{139.639}$ & $\mathbf{369.750}$ & $\mathbf{455.500}$  & $\mathbf{539.750}$ & $\mathbf{100}$  \\
\cline{2-10}
& \multirow{ 4}{*}{nguyen4}
& $(1+1)$-GP & $35040.000$ & $0.000$ & $0.000$ & $35040.000$ & $35040.000$  & $35040.000$ & $1$  \\
& & ($1+\lambda$)-GP & $26001.000$ & $0.000$ & $0.000$ & $26001.000$ & $26001.000$  & $26001.000$ & $1$  \\
& & Canonical-GP & $38804.500$ & $13591.931$ & $2774.441$ & $30006.500$ & $38099.000$  & $45942.500$ & $25$  \\
& & \textbf{FTG} & $\mathbf{713.710}$ & $\mathbf{1046.524}$ & $\mathbf{104.652}$ & $\mathbf{380.000}$ & $\mathbf{472.500}$  & $\mathbf{547.000}$ & $\mathbf{100}$  \\
\cline{2-10}
& \multirow{ 4}{*}{nguyen5}
& $(1+1)$-GP & $49371.571$ & $16847.311$ & $6367.685$ & $44891.000$ & $47935.000$  & $55453.500$ & $7$  \\
& & ($1+\lambda$)-GP & $65723.222$ & $16784.767$ & $5594.922$ & $49501.000$ & $63001.000$  & $74001.000$ & $9$  \\
& & Canonical-GP & $69099.500$ & $32249.826$ & $16124.913$ & $61131.500$ & $83666.000$  & $91634.000$ & $4$  \\
& & \textbf{FTG} & $\mathbf{427.800}$ & $\mathbf{699.765}$ & $\mathbf{69.977}$ & $\mathbf{233.250}$ & $\mathbf{333.500}$  & $\mathbf{417.750}$ & $\mathbf{100}$  \\
\cline{2-10}
& \multirow{ 4}{*}{nguyen6}
& $(1+1)$-GP & $19330.556$ & $25227.111$ & $8409.037$ & $1443.000$ & $4470.000$  & $31818.000$ & $9$  \\
& & ($1+\lambda$)-GP & $40101.000$ & $41206.068$ & $13030.503$ & $2501.000$ & $19501.000$  & $85626.000$ & $10$  \\
& & Canonical-GP & $20753.824$ & $14834.565$ & $1555.085$ & $10958.000$ & $16934.000$  & $24902.000$ & $91$  \\
& & \textbf{FTG} & $\mathbf{580.120}$ & $\mathbf{883.274}$ & $\mathbf{88.327}$ & $\mathbf{339.500}$ & $\mathbf{407.500}$  & $\mathbf{468.250}$ & $\mathbf{100}$  \\
\cline{2-10}
& \multirow{ 4}{*}{nguyen7}
& $(1+1)$-GP & $66302.125$ & $23275.136$ & $5818.784$ & $47527.250$ & $69151.500$  & $87158.000$ & $16$  \\
& & ($1+\lambda$)-GP & $64334.333$ & $23194.228$ & $9469.004$ & $50876.000$ & $71001.000$  & $79876.000$ & $6$  \\
& & Canonical-GP & $87934.571$ & $10288.448$ & $3888.668$ & $84164.000$ & $87152.000$  & $97112.000$ & $7$  \\
& & \textbf{FTG} & $\mathbf{253.360}$ & $\mathbf{476.742}$ & $\mathbf{47.674}$ & $\mathbf{131.000}$ & $\mathbf{197.000}$  & $\mathbf{268.250}$ & $\mathbf{100}$  \\
\cline{2-10}
& \multirow{ 4}{*}{nguyen8}
& $(1+1)$-GP & $62259.625$ & $25861.750$ & $9143.510$ & $32621.000$ & $73068.500$  & $78242.250$ & $8$  \\
& & ($1+\lambda$)-GP & $57401.000$ & $20477.793$ & $9157.947$ & $41501.000$ & $56001.000$  & $69501.000$ & $5$  \\
& & Canonical-GP & $\infty$ & $0$ & $0$ & $\infty$ & $\infty$  & $\infty$ & $0$  \\
& & \textbf{FTG} & $\mathbf{263.870}$ & $\mathbf{563.843}$ & $\mathbf{56.384}$ & $\mathbf{143.000}$ & $\mathbf{194.500}$  & $\mathbf{280.250}$ & $\mathbf{100}$  \\

\hline

\multirow{36}{*}{\rotatebox{90}{$\sum\limits_{\BF{x} \in \X} \br{F(\BF{x}) - \Fest(\BF{x})}^2 < 10^{-4}$}}
& \multirow{ 4}{*}{koza1}
& $(1+1)$-GP & $39868.727$ & $29222.709$ & $8810.978$ & $22789.000$ & $30884.000$  & $48371.500$ & $11$  \\
& & ($1+\lambda$)-GP & $38477.190$ & $33223.800$ & $7250.027$ & $4501.000$ & $38501.000$  & $62501.000$ & $21$  \\
& & Canonical-GP & $23168.456$ & $11348.642$ & $1276.822$ & $15938.000$ & $19922.000$  & $26147.000$ & $79$  \\
& & \textbf{FTG} & $\mathbf{575.920}$ & $\mathbf{1873.869}$ & $\mathbf{187.387}$ & $\mathbf{203.000}$ & $\mathbf{326.500}$  & $\mathbf{419.250}$ & $\mathbf{100}$  \\
\cline{2-10}
& \multirow{ 4}{*}{koza2}
& $(1+1)$-GP & $44888.062$ & $26959.736$ & $3891.303$ & $22120.750$ & $42285.500$  & $71993.000$ & $48$  \\
& & ($1+\lambda$)-GP & $39501.000$ & $26580.068$ & $3877.101$ & $17501.000$ & $31501.000$  & $55501.000$ & $47$  \\
& & Canonical-GP & $75947.000$ & $14661.428$ & $4232.390$ & $62874.500$ & $82421.000$  & $88770.500$ & $12$  \\
& & \textbf{FTG} & $\mathbf{452.250}$ & $\mathbf{560.254}$ & $\mathbf{56.025}$ & $\mathbf{261.750}$ & $\mathbf{342.500}$  & $\mathbf{418.250}$ & $\mathbf{100}$  \\
\cline{2-10}
& \multirow{ 4}{*}{koza3}
& $(1+1)$-GP & $41516.389$ & $22617.865$ & $3769.644$ & $22595.250$ & $36121.500$  & $59148.000$ & $36$  \\
& & ($1+\lambda$)-GP & $52709.333$ & $23694.636$ & $4836.647$ & $34626.000$ & $50751.000$  & $72876.000$ & $24$  \\
& & Canonical-GP & $84662.000$ & $0.000$ & $0.000$ & $84662.000$ & $84662.000$  & $84662.000$ & $1$  \\
& & \textbf{FTG} & $\mathbf{338.780}$ & $\mathbf{403.315}$ & $\mathbf{40.331}$ & $\mathbf{213.500}$ & $\mathbf{290.500}$  & $\mathbf{372.000}$ & $\mathbf{100}$  \\
\cline{2-10}
& \multirow{ 4}{*}{nguyen3}
& $(1+1)$-GP & $56039.100$ & $22191.570$ & $7017.591$ & $38728.250$ & $60560.000$  & $74392.000$ & $10$  \\
& & ($1+\lambda$)-GP & $44151.000$ & $22150.677$ & $7004.659$ & $29376.000$ & $41251.000$  & $63376.000$ & $10$  \\
& & Canonical-GP & $29427.684$ & $15085.569$ & $1998.132$ & $20918.000$ & $24404.000$  & $34862.000$ & $58$  \\
& & \textbf{FTG} & $\mathbf{638.860}$ & $\mathbf{1356.932}$ & $\mathbf{135.693}$ & $\mathbf{292.750}$ & $\mathbf{378.000}$  & $\mathbf{458.250}$ & $\mathbf{100}$  \\
\cline{2-10}
& \multirow{ 4}{*}{nguyen4}
& $(1+1)$-GP & $61324.375$ & $25664.585$ & $9073.801$ & $43417.250$ & $67352.500$  & $79230.500$ & $8$  \\
& & ($1+\lambda$)-GP & $64376.000$ & $32329.118$ & $11430.069$ & $46501.000$ & $74501.000$  & $92126.000$ & $8$  \\
& & Canonical-GP & $38060.692$ & $13207.060$ & $2590.117$ & $29259.500$ & $35360.000$  & $46191.500$ & $27$  \\
& & \textbf{FTG} & $\mathbf{579.150}$ & $\mathbf{923.225}$ & $\mathbf{92.323}$ & $\mathbf{323.750}$ & $\mathbf{406.500}$  & $\mathbf{478.000}$ & $\mathbf{100}$  \\
\cline{2-10}
& \multirow{ 4}{*}{nguyen5}
& $(1+1)$-GP & $41297.265$ & $26776.483$ & $4592.129$ & $16645.750$ & $37397.000$  & $62263.750$ & $34$  \\
& & ($1+\lambda$)-GP & $52898.059$ & $24165.112$ & $4144.283$ & $36126.000$ & $48251.000$  & $72001.000$ & $34$  \\
& & Canonical-GP & $62058.333$ & $23981.192$ & $5652.421$ & $40713.500$ & $64991.000$  & $82172.000$ & $18$  \\
& & \textbf{FTG} & $\mathbf{339.840}$ & $\mathbf{704.076}$ & $\mathbf{70.408}$ & $\mathbf{147.750}$ & $\mathbf{228.500}$  & $\mathbf{308.250}$ & $\mathbf{100}$  \\
\cline{2-10}
& \multirow{ 4}{*}{nguyen6}
& $(1+1)$-GP & $43444.455$ & $31297.138$ & $6672.572$ & $12896.250$ & $47592.000$  & $62725.500$ & $22$  \\
& & ($1+\lambda$)-GP & $39889.889$ & $30473.830$ & $7182.751$ & $8501.000$ & $35251.000$  & $68876.000$ & $18$  \\
& & Canonical-GP & $20573.231$ & $14742.725$ & $1545.457$ & $10958.000$ & $16934.000$  & $24902.000$ & $91$  \\
& & \textbf{FTG} & $\mathbf{427.710}$ & $\mathbf{719.904}$ & $\mathbf{71.990}$ & $\mathbf{224.000}$ & $\mathbf{298.500}$  & $\mathbf{395.500}$ & $\mathbf{100}$  \\
\cline{2-10}
& \multirow{ 4}{*}{nguyen7}
& $(1+1)$-GP & $40736.955$ & $22733.572$ & $3427.215$ & $20757.250$ & $38977.000$  & $56326.750$ & $44$  \\
& & ($1+\lambda$)-GP & $53778.778$ & $23257.549$ & $4475.917$ & $40001.000$ & $50001.000$  & $72751.000$ & $27$  \\
& & Canonical-GP & $74951.000$ & $13773.437$ & $2434.823$ & $66858.500$ & $75698.000$  & $82794.500$ & $32$  \\
& & \textbf{FTG} & $\mathbf{127.630}$ & $\mathbf{60.804}$ & $\mathbf{6.080}$ & $\mathbf{77.000}$ & $\mathbf{118.500}$  & $\mathbf{166.750}$ & $\mathbf{100}$  \\
\cline{2-10}
& \multirow{ 4}{*}{nguyen8}
& $(1+1)$-GP & $40714.536$ & $23084.951$ & $4362.646$ & $22117.500$ & $34446.000$  & $51288.000$ & $28$  \\
& & ($1+\lambda$)-GP & $58615.286$ & $23377.060$ & $3951.444$ & $40501.000$ & $57001.000$  & $74001.000$ & $35$  \\
& & Canonical-GP & $94622.000$ & $0.000$ & $0.000$ & $94622.000$ & $94622.000$  & $94622.000$ & $1$  \\
& & \textbf{FTG} & $\mathbf{141.650}$ & $\mathbf{99.291}$ & $\mathbf{9.929}$ & $\mathbf{93.250}$ & $\mathbf{122.000}$  & $\mathbf{166.250}$ & $\mathbf{100}$  \\

\hline

\multirow{36}{*}{\rotatebox{90}{$\sum\limits_{\BF{x} \in \X} \br{F(\BF{x}) - \Fest(\BF{x})}^2 < 10^{-3}$}}
& \multirow{ 4}{*}{koza1}
& $(1+1)$-GP & $34960.027$ & $25795.164$ & $4240.699$ & $13928.000$ & $30431.000$  & $55653.000$ & $37$  \\
& & ($1+\lambda$)-GP & $34178.083$ & $24313.810$ & $3509.396$ & $16126.000$ & $29751.000$  & $45626.000$ & $48$  \\
& & Canonical-GP & $27544.795$ & $19203.015$ & $2047.048$ & $16809.500$ & $20918.000$  & $31874.000$ & $88$  \\
& & \textbf{FTG} & $\mathbf{437.430}$ & $\mathbf{1836.204}$ & $\mathbf{183.620}$ & $\mathbf{144.500}$ & $\mathbf{224.500}$  & $\mathbf{306.500}$ & $\mathbf{100}$  \\
\cline{2-10}
& \multirow{ 4}{*}{koza2}
& $(1+1)$-GP & $23580.146$ & $21253.018$ & $2252.815$ & $6882.000$ & $17679.000$  & $32592.000$ & $89$  \\
& & ($1+\lambda$)-GP & $26989.095$ & $26172.843$ & $2855.691$ & $6376.000$ & $14001.000$  & $42751.000$ & $84$  \\
& & Canonical-GP & $66895.118$ & $19366.608$ & $3321.346$ & $55155.500$ & $70967.000$  & $83043.500$ & $34$  \\
& & \textbf{FTG} & $\mathbf{246.150}$ & $\mathbf{136.572}$ & $\mathbf{13.657}$ & $\mathbf{170.000}$ & $\mathbf{229.000}$  & $\mathbf{295.750}$ & $\mathbf{100}$  \\
\cline{2-10}
& \multirow{ 4}{*}{koza3}
& $(1+1)$-GP & $25311.600$ & $21484.742$ & $2773.668$ & $11815.750$ & $18703.500$  & $32022.000$ & $60$  \\
& & ($1+\lambda$)-GP & $36746.455$ & $22956.653$ & $3095.475$ & $18751.000$ & $30001.000$  & $52751.000$ & $55$  \\
& & Canonical-GP & $61056.800$ & $20552.624$ & $6499.310$ & $49428.500$ & $64244.000$  & $71340.500$ & $10$  \\
& & \textbf{FTG} & $\mathbf{175.450}$ & $\mathbf{71.727}$ & $\mathbf{7.173}$ & $\mathbf{124.750}$ & $\mathbf{171.500}$  & $\mathbf{228.000}$ & $\mathbf{100}$  \\
\cline{2-10}
& \multirow{ 4}{*}{nguyen3}
& $(1+1)$-GP & $44645.500$ & $27150.987$ & $4292.948$ & $22532.000$ & $38343.000$  & $63877.250$ & $40$  \\
& & ($1+\lambda$)-GP & $53448.368$ & $29194.175$ & $4735.921$ & $23626.000$ & $56501.000$  & $79126.000$ & $38$  \\
& & Canonical-GP & $34321.314$ & $17234.324$ & $2059.896$ & $21540.500$ & $27641.000$  & $39219.500$ & $71$  \\
& & \textbf{FTG} & $\mathbf{366.130}$ & $\mathbf{496.637}$ & $\mathbf{49.664}$ & $\mathbf{196.250}$ & $\mathbf{286.000}$  & $\mathbf{368.250}$ & $\mathbf{100}$  \\
\cline{2-10}
& \multirow{ 4}{*}{nguyen4}
& $(1+1)$-GP & $56991.758$ & $27871.068$ & $4851.730$ & $37894.000$ & $60825.000$  & $84615.000$ & $33$  \\
& & ($1+\lambda$)-GP & $53398.059$ & $26196.593$ & $4492.679$ & $33251.000$ & $49251.000$  & $76501.000$ & $34$  \\
& & Canonical-GP & $46888.700$ & $18754.362$ & $2965.325$ & $32372.000$ & $44822.000$  & $57396.500$ & $41$  \\
& & \textbf{FTG} & $\mathbf{384.200}$ & $\mathbf{533.676}$ & $\mathbf{53.368}$ & $\mathbf{222.500}$ & $\mathbf{294.000}$  & $\mathbf{378.000}$ & $\mathbf{100}$  \\
\cline{2-10}
& \multirow{ 4}{*}{nguyen5}
& $(1+1)$-GP & $26272.082$ & $21288.294$ & $2491.606$ & $10518.000$ & $21358.000$  & $39955.000$ & $73$  \\
& & ($1+\lambda$)-GP & $32882.944$ & $22147.625$ & $2610.123$ & $16876.000$ & $28751.000$  & $41126.000$ & $72$  \\
& & Canonical-GP & $51839.273$ & $20684.534$ & $3118.311$ & $34737.500$ & $49802.000$  & $65489.000$ & $44$  \\
& & \textbf{FTG} & $\mathbf{155.060}$ & $\mathbf{88.814}$ & $\mathbf{8.881}$ & $\mathbf{97.750}$ & $\mathbf{142.000}$  & $\mathbf{196.750}$ & $\mathbf{100}$  \\
\cline{2-10}
& \multirow{ 4}{*}{nguyen6}
& $(1+1)$-GP & $31346.864$ & $22724.749$ & $2958.510$ & $13997.000$ & $27941.000$  & $43901.000$ & $59$  \\
& & ($1+\lambda$)-GP & $38936.185$ & $24646.397$ & $3353.950$ & $21001.000$ & $34751.000$  & $59501.000$ & $54$  \\
& & Canonical-GP & $20425.413$ & $15019.988$ & $1565.942$ & $10958.000$ & $16187.000$  & $23906.000$ & $92$  \\
& & \textbf{FTG} & $\mathbf{241.360}$ & $\mathbf{370.258}$ & $\mathbf{37.026}$ & $\mathbf{141.750}$ & $\mathbf{192.500}$  & $\mathbf{238.250}$ & $\mathbf{100}$  \\
\cline{2-10}
& \multirow{ 4}{*}{nguyen7}
& $(1+1)$-GP & $24580.676$ & $23009.480$ & $2730.723$ & $7569.500$ & $15237.000$  & $32868.000$ & $71$  \\
& & ($1+\lambda$)-GP & $37465.286$ & $22246.560$ & $2658.973$ & $21626.000$ & $30751.000$  & $48126.000$ & $70$  \\
& & Canonical-GP & $57247.707$ & $17009.862$ & $1878.425$ & $43826.000$ & $55529.000$  & $72585.500$ & $82$  \\
& & \textbf{FTG} & $\mathbf{86.510}$ & $\mathbf{38.858}$ & $\mathbf{3.886}$ & $\mathbf{59.000}$ & $\mathbf{77.000}$  & $\mathbf{104.000}$ & $\mathbf{100}$  \\
\cline{2-10}
& \multirow{ 4}{*}{nguyen8}
& $(1+1)$-GP & $28216.351$ & $19159.528$ & $2537.742$ & $13659.000$ & $24183.000$  & $37461.000$ & $56$  \\
& & ($1+\lambda$)-GP & $41039.462$ & $23409.615$ & $2903.605$ & $21001.000$ & $35001.000$  & $54501.000$ & $65$  \\
& & Canonical-GP & $81460.571$ & $12561.500$ & $4747.801$ & $76943.000$ & $81176.000$  & $89891.000$ & $7$  \\
& & \textbf{FTG} & $\mathbf{106.290}$ & $\mathbf{54.586}$ & $\mathbf{5.459}$ & $\mathbf{60.750}$ & $\mathbf{96.000}$  & $\mathbf{144.000}$ & $\mathbf{100}$  \\
\bottomrule
\end{tabular}
}

\end{tabular}
\label{tab:rst1}
\end{table*}

\end{document}